\documentclass[lettersize,journal]{IEEEtran}
\usepackage{amsmath,amsfonts}
\usepackage{algorithmic}
\usepackage{algorithm}
\usepackage{array}
\usepackage[caption=false,font=normalsize,labelfont=sf,textfont=sf]{subfig}
\usepackage{textcomp}
\usepackage{stfloats}
\usepackage{url}
\usepackage{authblk}
\usepackage{verbatim}
\usepackage{graphicx}
\usepackage{cite}
\usepackage{multirow}
\usepackage{booktabs}
\usepackage{bbding}
\usepackage{color}
\usepackage{bbm}

\begin{document}

\title{DPNet: Dual-Path Network for Real-time Object Detection with Lightweight Attention}

\author{Quan Zhou,~\IEEEmembership{Member,~IEEE,} Huimin Shi, Weikang Xiang, Bin Kang, Xiaofu Wu, and Longin Jan Latecki
\thanks{Manuscript received XXXX XX, 2022; revised XXXX XX, 2022; accepted XXXX XX, 2022. This work was jointly supported in part by the National Natural Science Foundation of China under Grants 61876093, and Postgraduate Research $\&$ Practice Innovation Program of Jiangsu Province under Grants SJCX21$\_$0266.}
\thanks{\emph{Corresponding author: Quan Zhou.}}
\thanks{Quan Zhou, Huimin Shi, Weikang Xiang, and Xiaofu Wu are with National Engineering Research Center of Communications and Networking, Nanjing University of Posts \& Telecommunications, Nanjing, Jiangsu, P. R. China. (e-mail: \{quan.zhou, huimin.shi, weikang.xiang, xfuwu\}@njupt.edu.cn)}
\thanks{Bin Kang is with the Department of Internet of Things, Nanjing University of Posts \& Telecommunications, Nanjing, Jiangsu, P. R. China. (e-mail: kangbin@njupt.edu.cn)}
\thanks{Longin Jan Latecki is with the Department of Computer and Information Sciences, Temple University, Philadelphia, Pennsylvania, USA. (e-mail: latecki@temple.edu)}
}

\markboth{Journal of \LaTeX\ Class Files,~Vol.~14, No.~8, August~2021}%
{Shell \MakeLowercase{\textit{et al.}}: A Sample Article Using IEEEtran.cls for IEEE Journals}


\maketitle

\begin{abstract}
The recent advances of compressing high-accuracy convolution neural networks (CNNs) have witnessed remarkable progress for real-time object detection. To accelerate detection speed, lightweight detectors always have few convolution layers using single-path backbone. Single-path architecture, however, involves continuous pooling and downsampling operations, always resulting in coarse and inaccurate feature maps that are disadvantageous to locate objects. On the other hand, due to limited network capacity, recent lightweight networks are often weak in representing large scale visual data. To address these problems, this paper presents a dual-path network, named DPNet, with a lightweight attention scheme for real-time object detection. The dual-path architecture enables us to parallelly extract high-level semantic features and low-level object details. Although DPNet has nearly duplicated shape with respect to single-path detectors, the computational costs and model size are not significantly increased. To enhance representation capability, a lightweight self-correlation module (LSCM) is designed to capture global interactions, with only few computational overheads and network parameters. In neck, LSCM is extended into a lightweight cross-correlation module (LCCM), capturing mutual dependencies among neighboring scale features. We have conducted exhaustive experiments on MS COCO and Pascal VOC 2007 datasets. The experimental results demonstrate that DPNet achieves state-of-the-art trade-off between detection accuracy and implementation efficiency. Specifically, DPNet achieves 30.5\% AP on MS COCO test-dev and 81.5\% mAP on Pascal VOC 2007 test set, together with nearly 2.5M model size, 1.04 GFLOPs, and 164 FPS and 196 FPS for $320\times320$ input images of two datasets. 
\end{abstract}


\section{Introduction}

\IEEEPARstart{O}{bject} detection is a fundamental and challenging task in the field of computer vision. It aims to detect the minimum bounding boxes that cover objects of interest in input images, and assign associated semantic labels synchronously. Typically, the recent approaches based on convolutional neural networks (CNNs) can be roughly divided into two-stage \cite{fasterrcnn,he2017mask} and one-stage \cite{redmon2016you,liu2016ssd,redmon2018yolov3} detectors. The first category produces candidate boxes using region proposal networks at beginning, which will be subsequently refined in the next stage. Hence these detectors are always not efficient due to their multi-stage nature. In contrast, one-stage detectors \cite{redmon2016you,liu2016ssd,redmon2018yolov3} directly predict object categories and regress bounding boxes on convolutional feature maps. Since the whole pipeline is simplified, one-stage detectors always achieve faster inference speed than two-stage detectors \cite{fasterrcnn,he2017mask}. In spite of achieving remarkable progress, the vast majority of CNN-based detectors involve hundreds even thousands of convolutional layers and feature channels \cite{resnet,vgg}, where model size and implementation efficiency are unacceptable for real-world applications that require on-line estimations and real-time predictions, such as self-driving \cite{geiger2012we}, robot vision \cite{o2018evaluating}, and virtual reality \cite{liu2019edge}. 
\begin{figure}[t] 
\centering 
\includegraphics[width=0.48\textwidth]{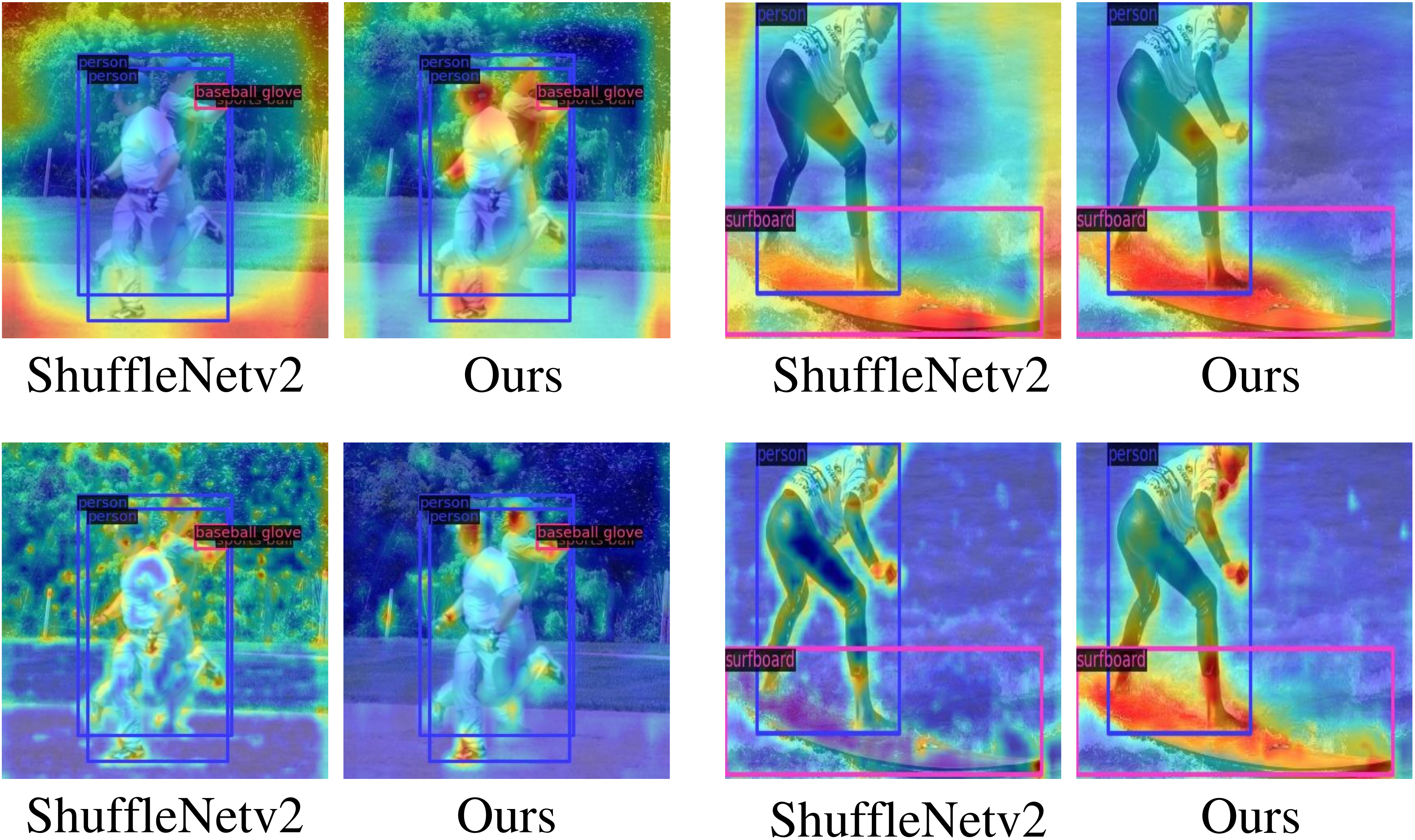} 
\caption{The feature heatmaps of two visual examples in MS COCO \cite{lin2014microsoft} validation set using ShuffleNetV2\cite{ma2018shufflenet} and DPNet. The red color denotes high responses, while blue color indicates low activations. For clarity, bounding boxes and associated labels have also been superimposed on objects of interest. Two rows show the heatmaps from low-resolution (deep layer) and high-resolution (shallow layer) features, respectively. Compared with ShuffleNetV2\cite{ma2018shufflenet}, the heatmaps of DPNet are more accurate, as most pixels with higher activations are located in object regions. (Best viewed in color)} 
\label{Fig:vis} 
\end{figure}

\begin{figure*}[ht] 
\centering 
\includegraphics[width=0.95\textwidth]{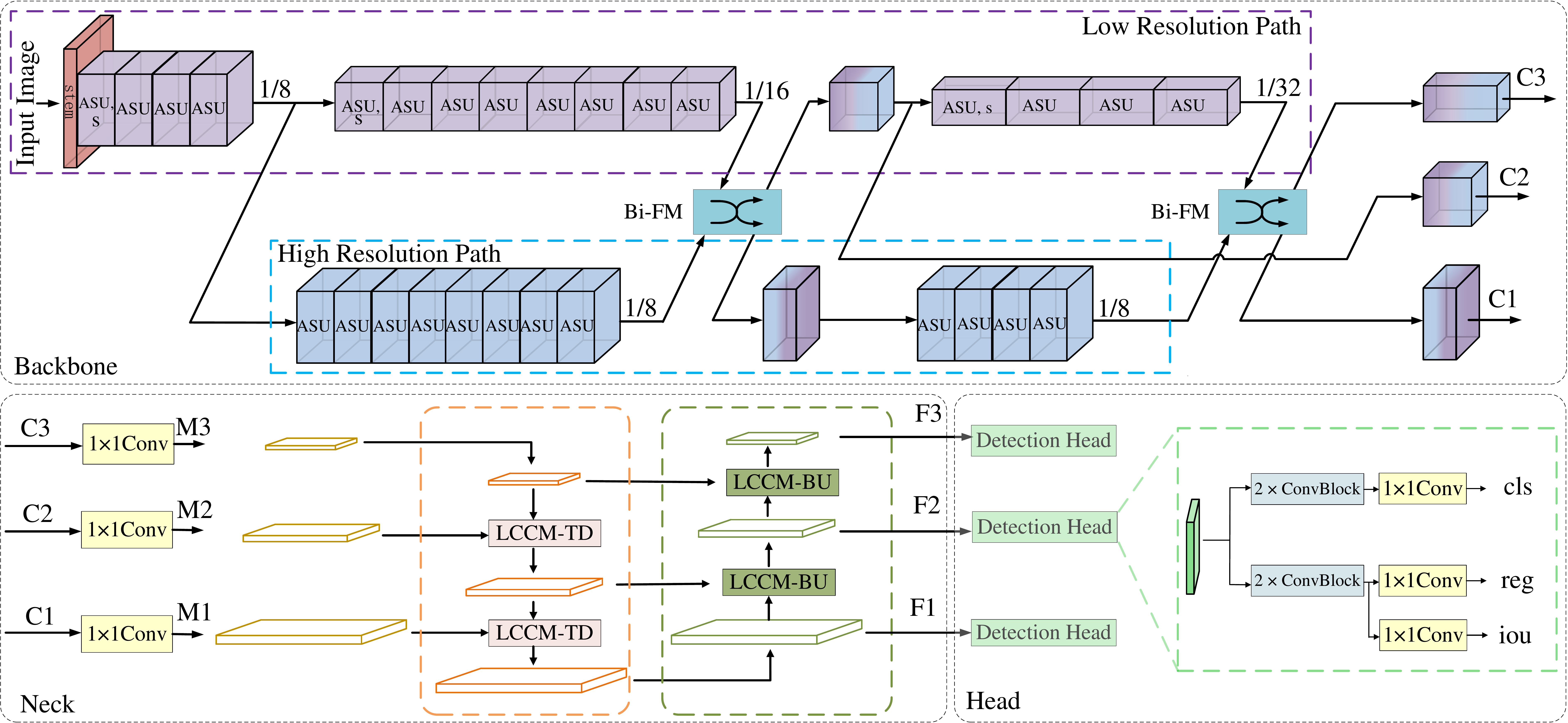} 
\caption{Overview architecture of DPNet. The backbone is mainly constructed by ASUs, together with a stem and two Bi-FMs. Meanwhile, it has dual-resolution structure: HRP and LRP, denoted by purple and blue dash rectangles, respectively. In neck, LCCM works in a bi-directional fashion, denoted by orange and green dash rectangles, to enhance cross-scale interactions. Our detection head uses several lightweight ConvBlocks for final predictions. (Best viewed in color)} 
\label{Fig:network} 
\end{figure*}

In order to adapt to real-world scenarios, a vast number of lightweight networks \cite{howard2017mobilenets, sandler2018mobilenetv2,qin2019thundernet} have been proposed for real-time object detection. Derived from \cite{howard2017mobilenets, sandler2018mobilenetv2, ma2018shufflenet} used for image classification, these lightweight networks prefer to directly inherit single-path architecture using lightweight convolution in their backbones. For instance, MobileNet-SSD \cite{howard2017mobilenets,sandler2018mobilenetv2} combines MobileNet with SSD-head. ThunerNet \cite{qin2019thundernet} adopts ShuffleNetV2 \cite{ma2018shufflenet} as backbone by replacing $3\times3$ depth-wise convolution with $5\times5$ depth-wise convolution. Pelee \cite{Pelee} employs lightweight backbone with dense structure, reducing output scales of SSD-head to save computational costs. Tiny-DSOD \cite{li2018tiny} introduces depth-wise convolutions both in backbone and feature pyramid network (FPN). Tiny-YOLO families \cite{redmon2017yolo9000,redmon2018yolov3,wang2021scaled} reduce the number of convolution layers or remove multi-scale outputs in neck. Although these advanced and efficient networks have achieved impressive detection results, they inherently suffer from following limitations:

\begin{itemize}
	\item Adopting aggressive down-sampling strategy (e.g., pooling and stride convolution), single-path architecture has been dominant in backbone design for real-time object detection \cite{sandler2018mobilenetv2,zhang2018shufflenet,wang2021scaled}. As fine object details are discarded step-by-step from shallow to deep convolutional layers, the produced high-level features are not beneficial to accurately locate objects. Two visual examples are given in Fig. \ref{Fig:vis}. The first row shows that ShuffleNetV2 \cite{ma2018shufflenet} prefers to extract features from surrounding areas of input images. Although lightweight detectors follow high-accuracy CNNs that employ FPN to relieve this problem \cite{li2018tiny}, simply integrating such inaccurate features from shallow to deep layers via element-wise addition or concatenation may be harmful for detecting objects \cite{li2020semantic}.
	\item Due to the limited network capacity, recent lightweight detectors may have weak representation ability of visual data \cite{mehta2021mobilevit}. In the second row of Fig. \ref{Fig:vis}, for example, high filter responses sometimes spread over clutter background (e.g., trees and sea), while areas containing the objects of interest are less activated. The underlying reason mainly lies in that, due to the limited receptive field, lightweight convolutions are very limited in encoding global dependencies \cite{mehta2021mobilevit}. Some networks prefer to utilize large convolution kernels (e.g., $31 \times 31$) \cite{peng2017large,ding2022scaling} or self-attention \cite{wang2018non}, yet they always involve huge computational cost and a heavy model size that is not suitable for real-time object detection. As a result, how to enhance feature representation ability with a small computational budget still remains unclear for lightweight object detection.
\end{itemize}

To address these shortcomings, this paper describes a dual-path network, called DPNet, with lightweight attention design for real-time object detection. As shown in Fig. \ref{Fig:network}, DPNet is composed of three components: backbone, neck, and detection head. To remedy the problem of abandoning object details, unlike previous lightweight detection networks \cite{howard2017mobilenets,sandler2018mobilenetv2,ma2018shufflenet} that always employ single-path structure, DPNet adopts a parallel-path architecture, leading to a dual-resolution backbone. More specifically, the resolution of low-resolution path (LRP) is gradually reduced as usual, where high-level semantic cues are encoded. Conversely, the resolution of high-resolution path (HRP) remains unchanged, where low-level spatial details are extracted. Both paths are significant for lightweight object detection. Considering the complementary nature of two sub-networks, a bi-direction fusion module (Bi-FM) is constructed to enhance communications between two paths, facilitating information flow among variable-resolution features. Although the backbone of DPNet has nearly duplicated shape with respect to single-path architecture \cite{howard2017mobilenets,sandler2018mobilenetv2,ma2018shufflenet}, the computational complexity and network size are not significantly increased.

In order to improve the representation capability of entire DPNet, on the other hand, we develop ShuffleNetV2 unit \cite{ma2018shufflenet} with a lightweight self-correlation module (LSCM), producing an attention-based shuffle unit (ASU). Similar to \cite{woo2018cbam,hu2018genet}, LSCM also produces spatial-wise and channel-wise attention maps, respectively. However, LSCM adopts a structure that is mimic to self-attention, rather than employing global pooling that is weak to represent element-wise dependencies. Furthermore, instead of exploring pixel-to-pixel/channel-to-channel correlations \cite{wang2018non,fu2019dual} that involve heavy calculations, LSCM investigates pixel-to-region/channel-to-group-channel dependencies in low-dimensional embeddings, thus saving huge amount of computational costs, while still retaining a powerful representation ability. In Fig. \ref{Fig:network}, to make full use of features with different resolutions in neck, LSCM is further extended to a lightweight cross-correlation module (LCCM). LCCM works in a bi-directional fashion: top-down (LCCM-TD) and bottom-up (LCCM-BU). LCCM-TD introduces high-level semantics to guide low-level features. Conversely, LCCM-BU utilizes low-level details to refine high-level cues. As shown in Fig. \ref{Fig:vis}, since DPNet inherits the merits from dual-path backbone and lightweight attention design, pixels within target object regions (e.g., human heads, hands, and feet, etc) are correctly activated, either in high-resolution or low-resolution features. In short, the main contributions of this paper are three-fold:
\begin{itemize}
	\item In contrast to mainstream lightweight detectors that use single-path backbone, DPNet employs a dual-path architecture that extracts high-level semantics and maintains low-level details, synchronously. Moreover, two paths complement each other to further boost performance. 
	\item We design a lightweight attention-based module LSCM that leverages implementing efficiency and representation capability. LSCM is computationally cheap as its computational complexity is linear to input feature resolutions. Even so, it still achieves powerful representation ability by investigating global spatial and channel interactions. We also extend LSCM to LCCM in neck part, where correlated dependencies are well explored between neighboring scale features with different resolutions. 
	\item We test DPNet on two challenging datasets: MS COCO \cite{lin2014microsoft} and Pascal VOC 2007 \cite{everingham2010pascal}. Extensive experiments demonstrate that our method achieves a state-of-the-art trade-off in terms of detection accuracy and implementation efficiency. More specifically, DPNet achieves 30.5\% AP on MS COCO test-dev and 81.5\% mAP on Pascal VOC 2007 test set, respectively, together with only 2.5M model size, 1.04 GFLOPs, and 164 FPS and 196 FPS for $320 \times 320$ input images of two datasets.
\end{itemize}

The remainder of this paper is organized as follows. After a brief introduction of related work in Section \ref{sec:Relatedwork}, we elaborate on the details of our DPNet in Section \ref{sec:DPNet}. Experimental results are given in Section \ref{sec:Experimence}, and Section \ref{sec:Conclusion} provides concluding remarks and future work.

\section{Related Work}\label{sec:Relatedwork}

In order to adapt to real-time applications, a vast number of methods have been proposed to compress object detectors,
such as quantization \cite{wang2021learning}, pruning \cite{han2015learning}, knowledge distillation \cite{dai2021general}, and lightweight model design \cite{Pelee,squeezedet,tang2020lightdet}. As our method belongs to the last category, we briefly review related work in this direction.

\subsection{Real-time Object Detection with Lightweight Design}

Although many stat-of-the-art one-stage detectors \cite{redmon2016you,liu2016ssd,redmon2018yolov3} have achieved real-time inference speed, their model size and computational costs are still unacceptable for real-time applications. To alleviate such limitation, designing lightweight detectors have attracted great attention in recent years \cite{qin2019thundernet,Pelee,squeezedet,tang2020lightdet}. Generally, real‐time object detectors can be broadly divided into two categories: CNN-based \cite{squeezedet,li2018tiny,xiong2021mobiledets} and Transformer-based \cite{mehta2021mobilevit,chen2021mobile} lightweight networks. 

The first category often adopts compact convolutions, such as group-wise \cite{sandler2018mobilenetv2}, depth-wise \cite{howard2017mobilenets} and factorized convolution \cite{li2021micronet}, to construct their backbones. As a pioneer work, Sequeezedet\cite{squeezedet} combines SqueezeNet \cite{iandola2016squeezenet} with YOLO head \cite{redmon2016you}. MobileNet-SSD \cite{sandler2018mobilenetv2} equips MobileNet \cite{howard2017mobilenets} with SSD head \cite{liu2016ssd} to achieve satisfactory detection results. ThunderNet \cite{qin2019thundernet} employs ShuffleNetV2 \cite{ma2018shufflenet} as backbone and designs a spatial attention module to capture global context. LightDet \cite{tang2020lightdet} also develops \cite{ma2018shufflenet} using a lightweight detail-preserving module. Tiny-DSOD \cite{li2018tiny} proposes an efficient depth-wise dense block to replace origin block in DenseNet \cite{huang2017densely}. PeleeNet \cite{Pelee} introduces an efficient variant of \cite{huang2017densely} to obtain real-time predictions. YOLO families \cite{redmon2017yolo9000,redmon2018yolov3}, as the most popular and advanced one-stage detectors, are often shrunk to a tiny version \cite{wang2021scaled} that compresses model size by reducing convolution layers. MobileDets\cite{xiong2021mobiledets}, another lightweight neural architecture search detectors, achieves better latency and compatibility on various mobile platforms. 

Transformer models \cite{liu2021Swin,chen2021mobile,mehta2021mobilevit}, on the other hand, begin to show their potential for object detection in recent years. As transformers derive from self-attention scheme \cite{vaswani2017attention} that involves huge computations, researchers reduce model size by designing lightweight CNN-transformer hybrids \cite{mehta2021mobilevit,chen2021mobile}. For instance, MobileViT \cite{mehta2021mobilevit} still adopts single-path architecture, inserting transformer block into inverted bottleneck module \cite{sandler2018mobilenetv2} for real-time object detection. Conversely, Mobile-Former\cite{chen2021mobile}  provides a lightweight mixture network, where CNN-branch extracts local features, while transformer-branch investigates global cues. Meanwhile, the local and global features interact with each other to improve performance.

In contrast to most of aforementioned methods designed in a single-path manner, our DPNet adopts a dual-path architecture. LRP extracts high-level rich semantics as well as HRP abstracts low-level accurate details, both of which are essential for real-time object detection. Although dual-path architecture has been explored for lightweight semantic segmentation \cite{yu2018bisenet,yu2021bisenet}, to our best knowledge, only Mobile-Former \cite{chen2021mobile} employs dual-path architecture. However, it still involves heavy model size, and the resolutions of both paths are gradually reduced as usual. In contrast, DPNet is a lightweight detector, where the resolution is kept constant throughout the entire HRP.

\subsection{Visual Attention}

Due to the powerful capability of capturing global context, visual attention \cite{hu2018squeeze,ecanet,gao2019global,vaswani2017attention} has been widely employed in CNNs to develop object detection. These networks can be roughly divided into two categories: squeeze-attention \cite{hu2018squeeze,cao2019gcnet,woo2018cbam} and self-attention \cite{vaswani2017attention,wang2018non,liu2021Swin}. 

The first category highlights important feature channels and positions, known as channel-wise and spatial-wise attention, through network learning. For example, SENet \cite{hu2018squeeze} utilizes pooling operation to encode global context. GSNet \cite{gao2019global} explores second-order global pooling to reweight feature channels. Besides channel attention, some networks \cite{hu2018genet,woo2018cbam} advocate to learn spatial attention that captures global positional context. For instance, CBAM\cite{woo2018cbam} adopts average and max pooling in channel and spatial dimensions, respectively. ECANet\cite{ecanet} replaces fully-connected layers with 1-D convolutional layers. Although capturing context clues through global pooling is computationally efficient, it is still weak in representing element-wise interactions.


The second category captures global context by calculating the correlation matrix between each image element. Non-local network \cite{wang2018non}, as a pioneer, models pixel-to-pixel relationship, where each position is reweighted by all other positions. Instead of computing dense attention maps, CCNet\cite{huang2019ccnet} proposes two consecutive criss-cross attentions to reduce computational costs. GCNet\cite{cao2019gcnet}, however, considers that only reweighting feature channels is enough to capture global context. ANNet \cite{zhu2019asymmetric} proposes spatial pyramid pooling in a non-local network to accelerate inference speed. Although these advanced networks achieve powerful capability to capture global context, they still suffer from heavy computations.


Different from these approaches, our DPNet employs LSCM to leverage representation ability and computational efficiency. From the perspective of capturing global context clues, LSCM also investigates spatial and channel interactions that is mimic to self-attention. However, LSCM is computationally cheap as it explores pixel-to-region/channel-to-group-channel mutual correlations, instead of dense pixel-to-pixel/channel-to-channel dependencies widely used in self-attention.


\subsection{Multi-scale Feature Integration}

Directly feeding convolutional features to detection head always leads to poor performance \cite{Pelee,howard2017mobilenets,sandler2018mobilenetv2}, thus fusing multi-scale convolution features often plays a significant role for real-time object detection \cite{li2018tiny,redmon2017yolo9000,redmon2018yolov3}. Following high-accuracy detectors \cite{lin2017feature,liu2018path}, early attempts \cite{li2018tiny,qin2019thundernet,tang2020lightdet} employ FPN to perform feature integration in a top-down manner. More specifically, the low-resolution features are first upsampled, and then combined with high-resolution features by element-wise addition or concatenation. To save computational overheads, ThunderNet \cite{qin2019thundernet} and Tiny-YOLOV4 \cite{wang2021scaled} both fuse features using a tiny version of FPN that reduces the number of outputs. LightDet \cite{tang2020lightdet} replaces 3$\times$3 standard convolution with compact depth-wise separable convolution in FPN. In contrast to this top-down fashion, EfficientDet \cite{tan2020efficientdet} adopts additional bottom-up strategy that downsamples high-resolution features to integrate with low-resolution ones. 

In contrast to integrating multi-scale feature maps via simple addition or stacking, LCCM is designed in neck to encode correlated dependencies from neighboring scale features. In spite of adopting bi-directional fusion strategy that is similar to EfficientDet \cite{tan2020efficientdet}, LCCM requires very few computational overheads due to its lightweight design. 


\begin{table}
\small\tabcolsep 0.5pt \caption{The detailed architecture of backbone in DPNet. $s$ denotes stride 2, LRP and HRP indicate low-resolution and high-resolution paths, respectively.}
\renewcommand\arraystretch{1.5}
\begin{center}
\scalebox{0.85}{
\begin{tabular}{l|c|c|c|c}
\toprule
\textbf{Layer}      &\textbf{LRP}  &\textbf{HRP}       &\multicolumn{2}{c}{\textbf{\quad\quad Operation}} \\
\midrule
$\begin{array}{l}1\end{array}$         &$\begin{array}{l}160\times160\times24\end{array}$ & &\multicolumn{2}{c}{~~~ $3 \times 3$ Conv, s} \\	
$\begin{array}{l}2\end{array}$         &$\begin{array}{l}80\times80\times24\end{array}$ & &\multicolumn{2}{c}{~~~ $2 \times 2$ Maxpooling} \\
\midrule
$\begin{array}{l}3\\4-6\end{array}$    &$\begin{array}{l}40\times40\times128\\40\times40\times128\end{array}$ & &\multicolumn{2}{c}{$\begin{array}{l} \mathrm{ASU,s}\\ \left[\mathrm{ASU}\right] \times 3\end{array} $}\\
\midrule
$\begin{array}{l}7\\8-14 \\ 15 \end{array}$    &$\begin{array}{l}20\times20\times256\\20\times20\times256 \\  {20\times20\times256}  \end{array}$  &$\begin{array}{l}40\times40\times128\ \\{40\times40\times128} \\ {40\times40\times128}\end{array}$ &\multicolumn{2}{l}{$ \begin{array}{l|l} \mathrm{ASU,s}&  \mathrm{ASU} \\ \left[\mathrm{ASU}\right] \times 7 & \left[\mathrm{ASU}\right] \times 7  \\ \hline \multicolumn{2}{c}{\mathrm{Bi-FM}} \end{array} $}\\
\midrule
$\begin{array}{l}16\\17-19 \\ 20 \end{array}$    &$\begin{array}{l}10\times10\times512\\10\times10\times512 \\  {10\times10\times512}  \end{array}$  &$\begin{array}{l}40\times40\times128\ \\{40\times40\times128} \\ {40\times40\times128}\end{array}$ &\multicolumn{2}{l}{$ \begin{array}{l|l} \mathrm{ASU,s}&  \mathrm{ASU} \\ \left[\mathrm{ASU}\right] \times 3 & \left[\mathrm{ASU}\right] \times 3  \\ \hline \multicolumn{2}{c}{\mathrm{Bi-FM}} \end{array} $}\\
\bottomrule
\end{tabular}}
\end{center}
\label{tab:DPNet}
\end{table}
\begin{figure*}[t!] 
\centering 
\includegraphics[width=0.95\textwidth]{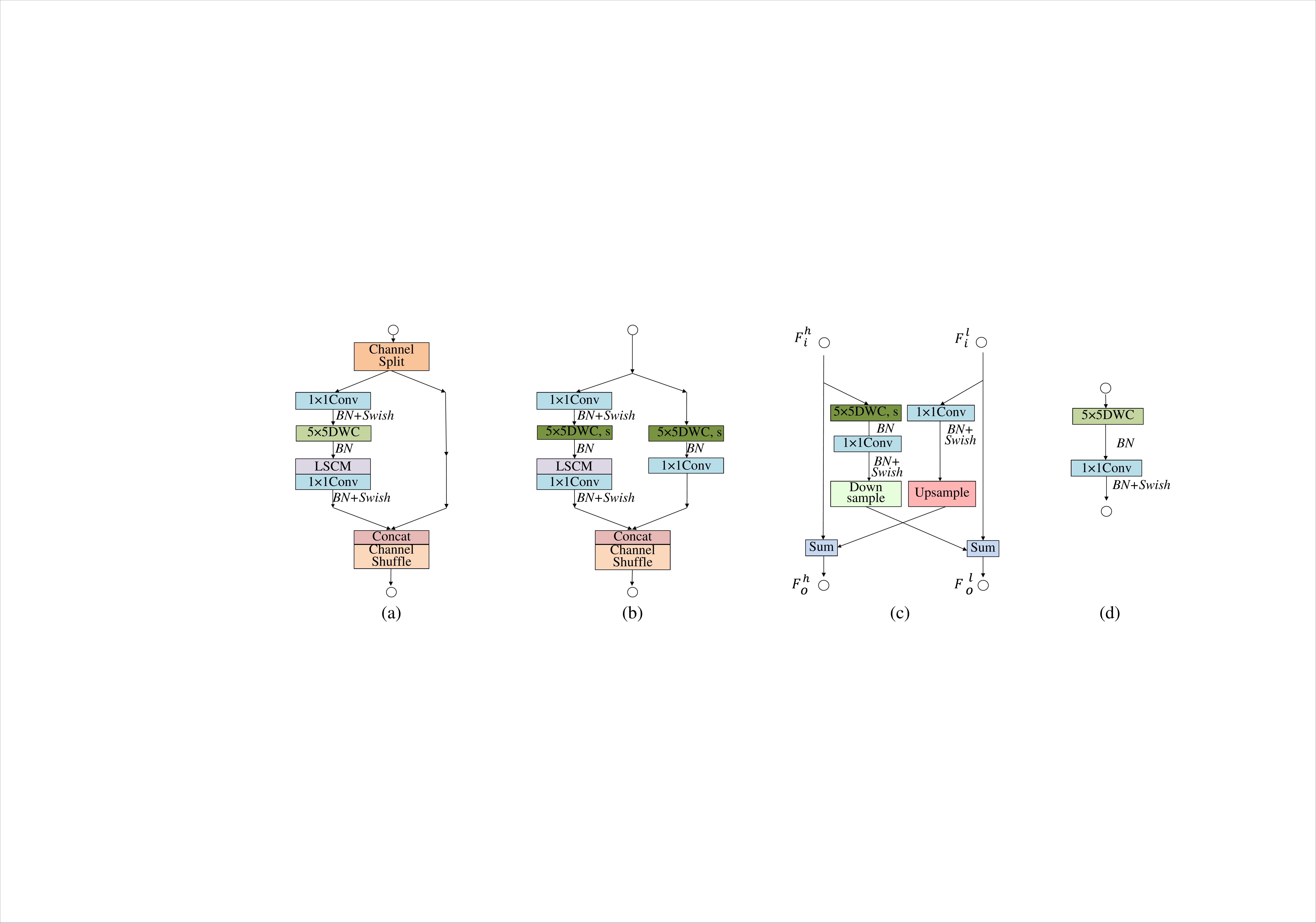} 
\caption{Overview of the units used in backbone and detection head. (a) ASU; and (b) stride version of ASU (s = 2); (c) Bi-FM and (d) ConvBlock. Conv stands for standard convolution, while DWC denotes depth-wise convolution. (Best viewed in color)} 
\label{Fig:Blocks} 
\end{figure*}

\section{Our Method}\label{sec:DPNet}

In this section, we first describe the entire lightweight dual-resolution architecture of DPNet, and then elaborate on the details of LSCM in backbone, together with LCCM in neck.

\subsection{DPNet}

The overall architecture of DPNet is depicted in Fig. \ref{Fig:network}. Concretely, our DPNet consists of three components: backbone, neck and detection head. Immediately below, we introduce each component in detail.

\subsubsection{Backbone}

The detailed structure of DPNet backbone is given in Table \ref{tab:DPNet}. More specifically, DPNet adopts a dual-resolution backbone, leading to a parallel-path architecture: LRP and HRP, respectively. Both paths are mainly constructed by a series of ASUs. Similar to traditional single-path detectors \cite{qin2019thundernet,Pelee,li2018tiny}, LRP employs a stem and multiple ASUs wtih stride 2, gradually producing convolution feature maps that have resolutions of $\frac{1}{2}$, $\frac{1}{4}$, $\frac{1}{8}$, $\frac{1}{16}$, and $\frac{1}{32}$ with respect to input image. Note the stem includes a stride 3$\times $3 convolution and a max-pooling, directly shrinking 4 times the input resolution. In order to obtain high-quality object details, on the other hand, HRP keeps a relatively high resolution of LRP, maintaining unchanged feature resolution that is $\frac{1}{8}$ of input image size. Among two paths, two Bi-FMs are employed to enhance cross-resolution feature integration and communication. Finally, as shown in Fig. \ref{Fig:network}, the combined features, denoted as $\lbrace$C1, C2, C3$\rbrace$ whose shapes are $\lbrace$40 $\times$40$\times$128, 20$\times$20$\times$256, 10$\times$10$\times$512$\rbrace$, serve as the multiple inputs to neck part, which helps to explore the mutual correlations. Next, we introduce ASU and Bi-FM in details, respectively.

\noindent \textbf{ASU.} 
As depicted in Fig. \ref{Fig:Blocks}(a), ASU adopts a split-transform-merge structure that leverages the residual connections and lightweight feature extractions. At the beginning of each ASU, input features are first split into two low-dimensional parts, transformed and identity branches, where each one has half channels of the input. The transformed branch serves as a residual function, while the identity branch is used to facilitate model training. Instead of using $3\times3$ depth-wise convolution \cite{ma2018shufflenet}, the transformed branch sequentially adopts depth-wise convolution with larger kernel size (e.g., $5\times5$) and the proposed LSCM, both which are used to obtain more powerful features. Thereafter, the outputs of two branches are merged using concatenation so that the number of channels keeps the same with respect to the input. Finally, feature channels are shuffled to enable information communication between two split branches. After the shuffle, the next ASU begins. Fig. \ref{Fig:Blocks}(b) also exhibits the stride version of ASU, used to reduce feature resolutions, where the $5 \times 5$ stride depth-wise convolutions are utilized in transformed and identity branches, respectively.

\noindent \textbf{Bi-FM.} Bi-FM serves as a bridge to enable communications between HRP and LRP in backbone. The detailed structure of Bi-FM is illustrated in Fig. \ref{Fig:Blocks}(c). Let $\textbf{\emph{F}}_{i}^{h} \in \mathbb{R}^{H \times W \times C}$ and $\textbf{\emph{F}}_{i}^{l} \in \mathbb{R}^{\frac{H}{n} \times \frac{W}{n} \times nC}, n \in \{2,4\}$, be inputs of Bi-FM, and $\textbf{\emph{F}}_{o}^{h} \in \mathbb{R}^{H \times W \times C}$ and $\textbf{\emph{F}}_{o}^{l} \in \mathbb{R}^{\frac{H}{n} \times \frac{W}{n} \times nC}$ be outputs of Bi-FM, respectively, where $H \times W$ stands for input resolution, and $C$ denotes channel number. More specifically, $\textbf{\emph{F}}_{i}^{l}$ first passes through an $1 \times 1$ convolution, and then upsampled with equal dimensions for following fusion with $\textbf{\emph{F}}_{i}^{h}$. On the other hand, to produce $\textbf{\emph{F}}_{o}^{l}$, $\textbf{\emph{F}}_{i}^{h}$ is fed into a $5\times5$ stride depth-wise convolution, and then downsampled with equal dimensions for next feature integration with $\textbf{\emph{F}}_{i}^{l}$.

\subsubsection{Neck}

Detection neck, also known as FPN \cite{lin2017feature}, is a fundamental component in state-of-the-art detectors to aggregate multi-scale features. Previous methods \cite{lin2017feature,liu2018path} utilize simple fusion strategy that employs bilinear interpolation and element-wise addition, often ignoring mutual dependencies across features with different resolutions. To this end, LCCM is adopted in the neck part of our DPNet, used to aggregate cross-resolution features from different convolution layers.

The detailed architecture of neck is illustrated in Fig. \ref{Fig:network}. Note LCCM works in a bi-directional fashion: top-down and bottom-up directions, denoted as LCCM-TD and LCCM-BU, respectively. LCCM-TD aims to extract high-level semantics for class identification, while LCCM-BU desires to strengthen low-level details for object localization. More specifically, receiving $\lbrace$C1, C2, C3$\rbrace$ produced from backbone as inputs, our neck begins at a series of $1 \times 1$ convolutions, producing features with equal channel numbers and various resolutions. These intermediate features, denoted as $\lbrace$M1, M2, M3$\rbrace$, are firstly fused in top-down path via two LCCM-TDs, and then aggregated in bottom-up path via two LCCM-BUs. Finally, the produced outputs, denoted as $\lbrace$F1, F2, F3$\rbrace$, where correlated interactions among neighboring scale feature maps are well integrated, are fed to lightweight detection head.



\subsubsection{Detection Head} 

Detection head learns projections that map features to final estimations. Some detection networks \cite{howard2017mobilenets,sandler2018mobilenetv2} employ lightweight backbones, yet involve SSD head \cite{liu2016ssd} that is too heavy to make predictions. The alternative approaches \cite{qin2019thundernet,li2018tiny,Pelee} design lightweight detection head to reduce model size. Similarly, our DPNet also adopts lightweight detection head to accelerate inference speed. As shown in Fig. \ref{Fig:Blocks}(d), instead of using 3$\times$3 depth-wise convolution \cite{qin2019thundernet,li2018tiny,Pelee}, DPNet utilizes compact convolutions with larger kernel size (e.g., $ 5\times5$) to enlarge receptive fields, increasing very limited model size. The detailed architecture of detection head is illustrated in Fig. \ref{Fig:network}. The input features $\lbrace$F1, F2, F3$\rbrace$, produced from neck part, undergo two successive ConvBlocks. Then, an 1 $\times$ 1 convolution is used to produce final outputs, receiving their supervision from the associated ground truth maps.

\subsection{LSCM and LCCM}


\subsubsection{LSCM}

The task of contextual formulation is to harvest surrounding information, which is always accomplished by global pooling \cite{hu2018genet,woo2018cbam,ecanet}. In spite of producing high-level features that represent entirety of an image, such networks are short in representation to provide element-wise interactions. Lots of alternative efforts \cite{vaswani2017attention,wang2018non,liu2021Swin} have been devoted to capture global context using dense attention maps, where the importance of each individual pixel is encoded by all other pixels. These methods, however, require large amount of computations. As the core unit of ASU, LSCM leverages computational efficiency and representation ability. Intuitively, there are two ways to save computational costs: shrinking number of elements and reducing feature dimensions. We thus introduce how LSCM works in these two aspects.

\begin{figure*}[t!] 
\centering 
\includegraphics[width=1.0\textwidth]{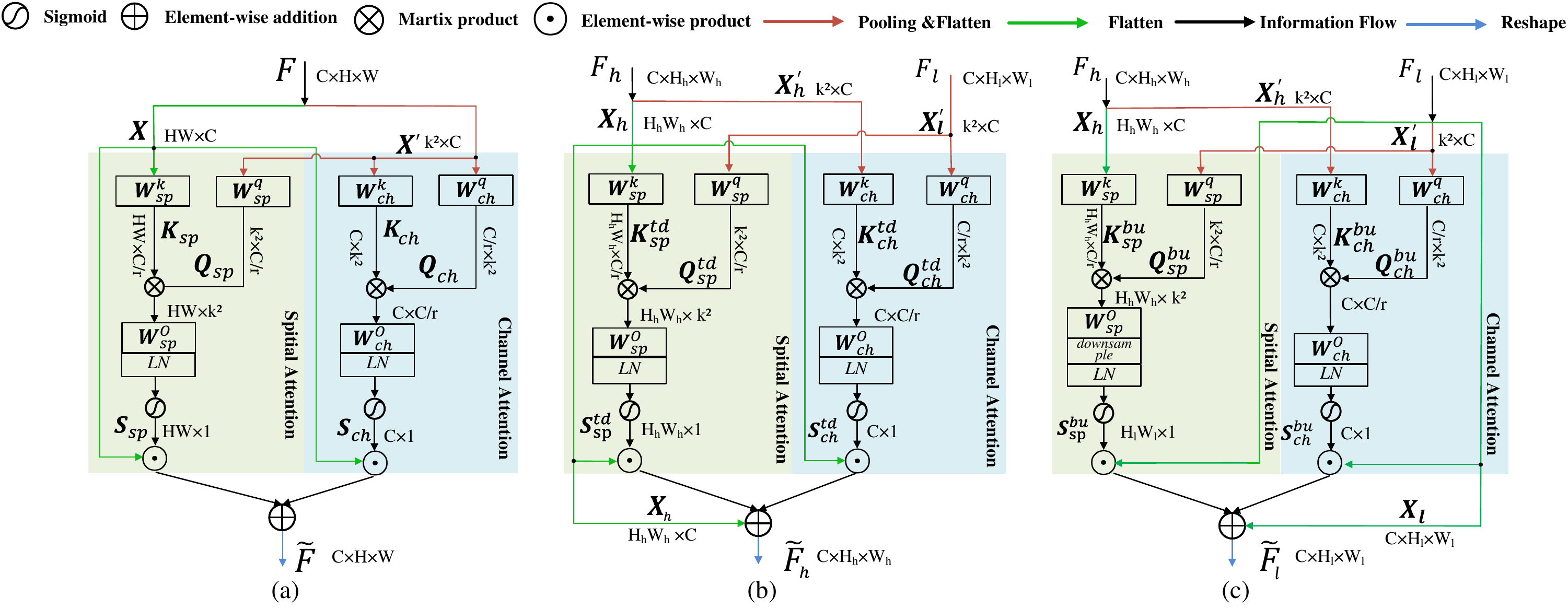} 
\caption{Overview of the lightweight attention employed in backbone and neck. (a) LSCM; (b) LCCM-TD, and (c) LCCM-BU. (Best viewed in color)} 
\label{Fig:LSCM} 
\end{figure*}

The detailed structure of LSCM is illustrated in Fig. \ref{Fig:LSCM}(a). Let $\textbf{\emph{F}}\in\mathbbm{R}^{C \times H \times W}$ be input features, where $W$, $H$, and $C$ stand for width, height, and channel number of inputs $\textbf{\emph{F}}$, respectively. To reduce image elements, we first apply pooling operation on input features $\textbf{\emph{F}}$, producing a compact representation $\textbf{\emph{R}}\in\mathbbm{R}^{C \times k \times k}, k^2 \ll W \times H$, where each element in $\textbf{\emph{R}}$ represents an image region that includes $WH/{k^2}$ pixels in $\textbf{\emph{F}}$. Then, both features $\textbf{\emph{F}}$ and $\textbf{\emph{R}}$ are flattened into two 2D sequences $\textbf{\emph{X}}\in\mathbbm{R}^{HW \times C}$ and $\textbf{\emph{X}}^{\prime}\in\mathbbm{R}^{k^{2} \times C}$, facilitating computations of following spatial and channel attention.


In spatial attention, two linear projections $\left\{\textbf{\emph{W}}_{sp}^k, \textbf{\emph{W}}_{sp}^q\right\}\in\mathbb{R}^{C \times C/r}$ are first learned to project input sequences $\textbf{\emph{X}}$ and $\textbf{\emph{X}}^{\prime}$ into two low-dimensional embeddings $\textbf{\emph{K}}_{sp}\in\mathbb{R}^{HW \times C/r}$ and $\textbf{\emph{Q}}_{sp}\in\mathbb{R}^{k^2 \times C/r}$, where $r$ is a non-negative scale factor that controls feature compression ratio:
\begin{equation}
    \textbf{\emph{K}}_{sp} = \textbf{\emph{X}}\textbf{\emph{W}}_{sp}^k \hspace{1cm} \textbf{\emph{Q}}_{sp}= \textbf{\emph{X}}^{\prime} \textbf{\emph{W}}_{sp}^q
\end{equation}

\noindent After that, the spatial pixel-to-region mutual correlations are calculated using a matrix product between $\textbf{\emph{K}}_{sp}$ and $\textbf{\emph{Q}}_{sp}$, which sequentially undergo a linear projection $\textbf{\emph{W}}_{sp}^O\in\mathbb{R}^{k^{2} \times 1}$, layer normalization $LN(\cdot)$, and sigmoid function $\sigma(\cdot)$, producing final spatial attention map $\textbf{\emph{S}}_{sp}\in\mathbbm{R}^{HW \times 1}$:
\begin{equation}
    \textbf{\emph{S}}_{sp}  = \sigma(LN(\textbf{\emph{K}}_{sp} \textbf{\emph{Q}}_{sp}^{\top}\textbf{\emph{W}}_{sp}^O))
\end{equation}

In channel attention, to reduce feature dimensions, a linear projection $\textbf{\emph{W}}_{ch}^q\in\mathbb{R}^{C/r \times C}$ is first learned to map input sequence $\textbf{\emph{X}}^{\prime}$ into a low-dimensional embedding $\textbf{\emph{Q}}_{ch}\in\mathbb{R}^{C/r \times k^2}$, where each channel in $\textbf{\emph{Q}}_{ch}$ represents a group of $r$ channels in $\textbf{\emph{X}}^{\prime}$. On the other hand, another linear projection $\textbf{\emph{W}}_{ch}^k\in\mathbb{R}^{C \times C}$ also maps input sequence $\textbf{\emph{X}}^{\prime}$ into $\textbf{\emph{K}}_{ch}$ $\in \mathbb{R}^{C \times k^2}$:

\begin{equation}
    \textbf{\emph{K}}_{ch} = \textbf{\emph{W}}_{ch}^k{\textbf{\emph{X}}^{\prime}}^{\top} \hspace{1cm} \textbf{\emph{Q}}_{ch} = \textbf{\emph{W}}_{ch}^q{\textbf{\emph{X}}^{\prime}}^{\top}
\end{equation}
Next, similar to spatial attention, the channel-to-group-channel correlations are computed using a matrix product between $\textbf{\emph{K}}_{ch}$ and $\textbf{\emph{Q}}_{ch}$, which sequentially undergo a linear projection $\textbf{\emph{W}}_{ch}^O$ $\in \mathbb{R}^{C/r \times 1}$, layer normalization $LN(\cdot)$, and sigmoid function $\sigma(\cdot)$, producing final channel attention map $\textbf{\emph{S}}_{ch}\in\mathbbm{R}^{C \times 1}$:

\begin{equation}\label{eq:LSCM_CA}
    \textbf{\emph{S}}_{ch}  = \sigma(LN(\textbf{\emph{K}}_{ch}\textbf{\emph{Q}}_{ch}^{\top}\textbf{\emph{W}}_{ch}^O))
\end{equation}
At the end, the learned spatial attention map $\textbf{\emph{S}}_{sp}$ and channel attention map $\textbf{\emph{S}}_{ch}$ are used to reweight input sequence $\textbf{\emph{X}}$, respectively, and thereafter combined using element-wise addition, producing integrated features $\widetilde{\textbf{\emph{X}}}\in\mathbb{R}^{HW \times C}$:

\begin{equation}
    \widetilde{\textbf{\emph{X}}} = (\textbf{\emph{S}}_{sp}\odot \textbf{\emph{X}}) \oplus
    (\textbf{\emph{S}}_{ch} \odot \textbf{\emph{X}}) 
\end{equation}
where $\oplus$ and $\odot$ are element-wise addition and multiplication, respectively. Note that two attention maps $\textbf{\emph{S}}_{sp}$ and $\textbf{\emph{S}}_{ch}$ are multiplied with input sequence $\textbf{\emph{X}}$ in the ways of column reweighting and row reweighting, respectively. The produced sequence $\widetilde{\textbf{\emph{X}}}$ is finally reshaped to $\widetilde{\textbf{\emph{F}}}\in\mathbb{R}^{C \times H \times W}$ with equal dimension with respect to input feature $\textbf{\emph{F}}$, which is ready for following 1$\times$1 convolution, as shown in Fig. \ref{Fig:Blocks}(a).


\begin{table}
\tabcolsep 0.1mm\caption{Computational complexity comparison between self-attention and the proposed LSCM. Herein, $n=W \times H$, and $c$ represents channel numbers of input features, respectively.}
	\begin{center}
	\begin{tabular}{c|c|c|c}
		\toprule 
		\multirow{1}{*}{Methods} & \multirow{1}{*}{Similarity} & \multirow{1}{*}{Reweight} & \multirow{1}{*}{Total}\\
		\midrule
		Non-local \cite{wang2018non} &$\mathcal{O}(n^{2}c)$  &$\mathcal{O}(n^{2}c)$ &$\mathcal{O}(2n^{2}c)$\\
		DANet \cite{fu2019dual} &$\mathcal{O}(n^{2}c+nc^{2})$  &$\mathcal{O}(n^{2}c+nc^{2})$  &$\mathcal{O}(2n^{2}c+2nc^{2})$\\
		\midrule
		LSCM &$\mathcal{O}(nc\frac{k^{2}}{r}+k^{2}\frac{c^2}{r})$  &$\mathcal{O}(2nc)$  &$\mathcal{O}(nc(2+\frac{k^{2}}{r})+k^{2}\frac{c^2}{r})$\\
		\bottomrule
	\end{tabular}\label{tab:complex}
	\end{center}
\end{table}

We also analyze the computational complexity of proposed LSCM, and compare it with recent self-attention networks \cite{wang2018non,fu2019dual}, as they all have powerful representation capability in investigating global dependencies. Table \ref{tab:complex} reports the comparison results that include spatial and channel attention together. For clarity, we only take spatial attention into account.
Previous methods \cite{wang2018non,fu2019dual} and our LSCM both involve two computational steps: calculating element-wise similarities and reweighting input features. In self-attention \cite{wang2018non}, computing dense spatial attention and rewrighting features both require $n^2c$ operations, leading to a quadric complexity of input resolution. On the contrary, our LSCM only needs $nc\frac{k^{2}}{r}$ operations that is linear to input resolution, as feature elements have been greatly reduced using global pooling. Furthermore, since the produced attention map is very simple, the reweighting process only requires $nc$ operations, instead of $n^2c$ in self-attention that is once again quadric to input feature size. 


\subsubsection{LCCM}

This section extends LSCM to a multiple inputs version, known as LCCM, since it is used in neck part to combine multi-scale features. Recalling that LCCM works in a bi-directional fashion: top-down and bottom-up, denoted as LCCM-TD and LCCM-BU, respectively. Since they work in a similar way, we only elaborate on LCCM-TD in this section, and then point out its major differences with LCCM-BU.

The detailed architecture of LCCM-TD is exhibited in Fig. \ref{Fig:LSCM}(b). Generally, LCCM-TD shares similar structure with LSCM, except two inputs that have different resolutions. Let $\textbf{\emph{F}}_h\in\mathbbm{R}^{C\times H_h \times W_h}$ and $\textbf{\emph{F}}_l\in\mathbbm{R}^{C\times H_l \times W_l}$ be high-resolution and low-resolution input features, respectively. Herein, $H_h=2H_l$, $W_h=2W_l$, as $\textbf{\emph{F}}_h$ and $\textbf{\emph{F}}_l$ come from neighboring scale convolution layers. In order to explore cross-layer interactions and save computational cost, the resolutions of $\textbf{\emph{F}}_h$ and $\textbf{\emph{F}}_l$ have to be shrunk at the same time using global pooling, and then flattened into two 2D sequences $\textbf{\emph{X}}_h^{\prime}\in\mathbbm{R}^{k^{2}\times C}$ and $\textbf{\emph{X}}_l^{\prime}\in\mathbbm{R}^{k^{2}\times C}$, respectively. Note $k^2\ll H_l\times W_l<H_h\times W_h$. Meanwhile, the input features $\textbf{\emph{F}}_h$ are also flattened into a 2D sequence $\textbf{\emph{X}}_h\in\mathbbm{R}^{H_h W_h\times C}$, ready to participate following computations of spatial and channel attention.


In spatial attention, input features ${\textbf{\emph{X}}_{h}}$ and $\textbf{\emph{X}}_l^{\prime}$ undergo two linear projections $\left\{\textbf{\emph{W}}_{sp}^k, \textbf{\emph{W}}_{sp}^q\right\}\in\mathbb{R}^{C \times C/r}$, resulting in two low-dimensional embeddings ${{\textbf{\emph{K}}}_{sp}^{td}}\in\mathbb{R}^{H_hW_h\times C/r}$ and ${\textbf{\emph{Q}}}_{sp}^{td}\in\mathbb{R}^{k^{2} \times C/r}$, respectively, where $r$ is a non-negative scale factor that controls feature compression ratio:


\begin{equation}
	\textbf{\emph{K}}_{sp}^{td}= \textbf{\emph{X}}_{h} \textbf{\emph{W}}_{sp}^k \hspace{1cm} {\textbf{\emph{Q}}}_{sp}^{td}  = \textbf{\emph{X}}_{l}^{\prime}\textbf{\emph{W}}_{sp}^q 
\end{equation}
After that, the spatial cross-layer interactions are calculated using a matrix product between $\textbf{\emph{K}}_{sp}^{td}$, and $\textbf{\emph{Q}}_{sp}^{td}$, which are sequentially fed into a linear projection $\textbf{\emph{W}}_{sp}^O\in\mathbb{R}^{k^{2} \times 1}$, layer normalization $LN(\cdot)$, and sigmoid function $\sigma(\cdot)$, producing final spatial attention map $\textbf{\emph{S}}_{sp}^{td}\in\mathbbm{R}^{H_hW_h \times 1}$:


\begin{equation}
    {\textbf{\emph{S}}}_{sp}^{td}  = \sigma(LN({\textbf{\emph{K}}}_{sp}^{td} {{\textbf{\emph{Q}}}_{sp}^{td}}^{\top}\textbf{\emph{W}}_{sp}^O))
\end{equation}

In channel attention, a linear projection $\textbf{\emph{W}}_{ch}^q\in\mathbb{R}^{C/r \times C}$ first maps input sequence $\textbf{\emph{X}}_l^{\prime}$ into a low-dimensional embedding $\textbf{\emph{Q}}_{ch}^{td}\in\mathbb{R}^{C/r \times k^2}$. Then, another linear projection $\textbf{\emph{W}}_{ch}^k\in\mathbb{R}^{C \times C}$ is learned to map input sequence $\textbf{\emph{X}}_h^{\prime}$ into $\textbf{\emph{K}}_{ch}^{td}\in \mathbb{R}^{C \times k^2}$:


\begin{equation}
    {\textbf{\emph{K}}}_{ch}^{td}= \textbf{\emph{W}}_{ch}^k{\textbf{\emph{X}}_h^{\prime}}^{\top} \hspace{1cm} {\textbf{\emph{Q}}}_{ch}^{td}  = \textbf{\emph{W}}_{ch}^q{\textbf{\emph{X}}_l^{\prime}}^{\top}
\end{equation}
Next, similar to spatial attention, the cross-layer correlations are computed using a matrix product between $\textbf{\emph{K}}_{ch}^{td}$ and $\textbf{\emph{Q}}_{ch}^{td}$, which sequentially undergo a linear projection $\textbf{\emph{W}}_{ch}^O$ $\in \mathbb{R}^{C/r \times 1}$, layer normalization $LN(\cdot)$, and sigmoid function $\sigma(\cdot)$, producing final channel attention map $\textbf{\emph{S}}_{ch}^{td}\in\mathbbm{R}^{C \times 1}$:

\begin{equation}
    {\textbf{\emph{S}}}_{ch}^{td}  = \sigma(LN({\textbf{\emph{K}}}_{ch}^{td} {{\textbf{\emph{Q}}}_{ch}^{td}}^{\top}\textbf{\emph{W}}_{ch}^O))
\end{equation}
Thereafter, the learned spatial attention map ${\textbf{\emph{S}}}_{ch}^{td}$ and channel attention map ${\textbf{\emph{S}}}_{sp}^{td}$ are used to reweight high-resolution input $\textbf{\emph{X}}_h$, respectively, and combined using element-wise addition, producing integrated features $\textbf{\emph{X}}_w\in\mathbb{R}^{H_hW_h \times C}$:

\begin{equation}\label{eq:td_weighting}
\textbf{\emph{X}}_w =({\textbf{\emph{S}}}_{sp}^{td}\odot \textbf{\emph{X}}_h) \oplus ({{\textbf{\emph{S}}}_{ch}^{td}} \odot \textbf{\emph{X}}_h)
\end{equation}
The entire reweighting process serves as a residual function that facilitates training LCCM-TD in an end-to-end manner:

\begin{equation}
	\widetilde{\textbf{\emph{X}}}_{h} =\textbf{\emph{X}}_w \oplus \textbf{\emph{X}}_h
\end{equation}
Note that two weighting operations in Eq. (\ref{eq:td_weighting}) are column and row reweighting, respectively, similar with LSCM. The produced sequence $\widetilde{\textbf{\emph{X}}}_h$ is finally reshaped to $\widetilde{\textbf{\emph{F}}}_h\in\mathbb{R}^{C \times H_h \times W_h}$ with equal dimension with respect to input feature $\textbf{\emph{F}}_h$, which is ready for next integration, as shown in Fig. \ref{Fig:network}.


Regarding to LCCM-BU, its detailed architecture is shown in Fig. \ref{Fig:LSCM}(c). There is only one difference with respect to LCCM-TD: When spatial attention is computed, the resolution has to be downsampled two times for exact reweighting and identity mapping.

\section{Experiments}\label{sec:Experimence}

In order to evaluate the proposed DPNet, we have conducted exhausted experiments on two challenging object detection datasets: MS COCO\cite{lin2014microsoft} and Pascal VOC 2007 \cite{everingham2010pascal}, including comprehensive comparisons with recent real-time detection networks and ablation studies. Experimental results show that, our DPNet achieves a state-of-the-art trade-off in terms of detection accuracy and implementing efficiency.

\subsection{ Dataset and Evaluation metrics}
\subsubsection{MS COCO}
MS COCO dataset \cite{lin2014microsoft} is the most popular object detection dataset in the field of computer vision. It involves 80 categories, containing 118K training, 5K validation and 20K testing images. As usual, we have conducted all experiments using training set to train our DPNet. A system-level comparison with recent state-of-the-art networks is reported on test-dev. Otherwise, a series of ablation studies are reported on validation set.
\subsubsection{Pascal VOC 2007}
Pascal VOC 2007 dataset \cite{everingham2010pascal} is a relative smaller dataset, compared with MS COCO \cite{lin2014microsoft}, only including 20 classes for object detection. Following \cite{Pelee,li2018tiny}, DPNet is trained on the union of VOC 2007 and VOC 2012 trainval set that contains 16,551 images together, and evaluated on the VOC 2007 test set, including 4,952 images. 

\begin{table*}[t!]
\tabcolsep 0.9mm \caption{Comparison with the high-accuracy and real-time object detectors in terms of detection accuracy and implementing efficiency on MS COCO test-dev \cite{lin2014microsoft}. ‘-’ denotes the results are not reported. ‘$\dagger$’ and ‘$\ddagger$’ mean DPNet is pre-trained using ImageNet 1K and 21K dataset \cite{deng2009imagenet}, respectively. Note the green and red numbers are with respect to the second-ranked method \cite{qin2019thundernet}.} 
	\begin{center}
	\begin{tabular}{l|c|c|c|c|c|c|c|c|c}
			\toprule 
			Method&Year&Backbone&Input Size&FLOPs (G) & Params (M) & $AP$ (\%) &$AP_{50}$ (\%) &$AP_{75}$ (\%) & FPS\\
			\midrule		
			YOLOV3 \cite{redmon2018yolov3}&Arxiv2018&DarkNet-53 & $320\times320$ &19.6 &62.3 &28.2 &51.5&29.7&56\\
			Mobile-Former-RetinaNet\cite{chen2021mobile} &CVPR2022 &Mobile-Former &$1333\times800$  &161 &14.4 &34.2&53.4&36.0&--\\
			RetainNet \cite{lin2017focal} &ICCV2017&ResNet-50& $1333\times800$ &251.3 &34.2 &35.7&55.0&38.5&19\\
			ATSS\cite{zhang2020bridging}  &CVPR2020&ResNet-50 & $1333\times800$&205 & 32 &43.5 &61.9&47.0 &28.3\\		
			Sparse R-CNN\cite{sun2021sparse}  &CVPR2021&ResNet-50 & $1333\times800$&109 & 53 &44.5 &63.5&48.2 &21.0\\
			Swin \cite{liu2021Swin}&ICCV2021 &Swin-Tiny &$1333\times800$ &245 &38.5 &45.5 &66.3&48.8&22.3\\
			YOLOV4 \cite{wang2021scaled} &CVPR2020&CSPDarkNet53 & $608\times608$&109 & 53 &45.5 &64.1&49.5 &62\\
			\midrule
			Tiny-YOLOV3 \cite{redmon2018yolov3} &Arxiv2018&Tiny-DarkNet & $416\times416$&2.78 & 8.7 &16.0&33.1&-- &368\\
			Tiny-YOLOV4 \cite{wang2021scaled} &CVPR2021& CSPDarkNet53-Tiny &$416\times416$&3.45 & 6.1 &21.7&40.2&--& $\textbf{371}$\\ 		
			MobileNetV2-SSDLite \cite{sandler2018mobilenetv2}  &CVPR2018 & MobileNetV2 &$320\times320$& \textbf{0.8} & 4.3 &21.3&35.4&21.8&69.9\\
			MobileNet-SSDLite \cite{sandler2018mobilenetv2}  &CVPR2018 & MobileNet &$320\times320$& 1.3 & -- &22.1&--&--&80\\
			Pelee \cite{Pelee} &NeurIPS2018 &PeleeNet&$304\times304$  &1.29 & 6.0 &22.4&38.3&22.9 &120\\
			Tiny-DSOD \cite{li2018tiny}&BMVC2018 &DDB-Net&$300\times300$  &1.12 & -- &23.2&40.4&22.8 & --\\
			Mobile-ViT-SSDLite \cite{mehta2021mobilevit}&ICLR2022 &Mobile-ViT-XS &$320\times320$ &--&2.7 &24.8&--&--&61\\
			MobileDets \cite{xiong2021mobiledets}&CVPR2021 & IBN+Fused+Tucker  &$320\times320$&1.43 & 4.85 &26.9&--&-- &-- \\
			Mobile-ViT-SSDLite \cite{mehta2021mobilevit}&ICLR2022	 &Mobile-ViT-S &$320\times320$  &--&5.7&27.7&--&--&80\\
			ThunderNet \cite{qin2019thundernet}&ICCV2019&SNet-535&$320\times320$&1.3 & -- &28.1 &46.2&29.6 & --\\			
			\midrule
			Ours &- &DPNet&$320\times320$&1.04$(\uparrow\textcolor{red}{\textbf{0.24}})$ & \textbf{2.5}$(\downarrow\textcolor{green}{\textbf{0.2}})$ &  \textbf{29.6}$(\uparrow\textcolor{red}{\textbf{1.5}})$ &$\textbf{46.9}(\uparrow\textcolor{red}{\textbf{0.7}})$  &$\textbf{31.3}(\uparrow\textcolor{red}{\textbf{1.7}})$ & 164\\
			Ours$^{\dagger}$ &- &DPNet&$320\times320$&1.04$(\uparrow\textcolor{red}{\textbf{0.24}})$ & \textbf{2.5}$(\downarrow\textcolor{green}{\textbf{0.2}})$ &  $\textbf{30.2}(\uparrow\textcolor{red}{\textbf{2.1}})$ &$\textbf{47.5}(\uparrow\textcolor{red}{\textbf{1.3}})$  &$\textbf{31.8}(\uparrow\textcolor{red}{\textbf{2.2}})$ & 164\\
			Ours$^{\ddagger}$ &- &DPNet&$320\times320$&1.04$(\uparrow\textcolor{red}{\textbf{0.24}})$ & \textbf{2.5}$(\downarrow\textcolor{green}{\textbf{0.2}})$ &  $\textbf{30.5}(\uparrow\textcolor{red}{\textbf{2.4}})$ &$\textbf{47.7}(\uparrow\textcolor{red}{\textbf{1.5}})$  &$\textbf{32.2}(\uparrow\textcolor{red}{\textbf{2.6}})$ & 164\\
			\bottomrule
	\end{tabular}
	\end{center}\label{tab:det_result}
\end{table*}

\subsubsection{Evaluation metrics}

For fair comparison with other state-of-the-art real-time detectors on MS COCO dataset \cite{lin2014microsoft}, we employ the standard evaluation metrics \cite{lin2017focal,redmon2018yolov3,wang2021scaled}, such as AP, AP$_{50}$, AP$_{75}$, AP$_{L}$, AP$_{M}$, and AP$_{S}$. More specifically, AP is averagely evaluated at IoU (intersection area over the union area between the predicted bounding boxes and ground-truth bounding boxes), ranged from 0.5 to 0.95 with updated step 0.05, reflecting the comprehensive performance of the detector. AP$_{50}$ and AP$_{75}$ are computed when IoU is 0.5 or 0.75, respectively. On the other hand, AP$_{L}$, AP$_{M}$, and AP$_{S}$ are used to evaluate the performance of bounding boxes whose areas are within the range of (0, $32^{2}]$, ($32^{2}$, $96^{2}$), [$96^{2}$, $+\infty$) pixels, representing the performance for small, medium, and large objects. For Pascal VOC 2007 \cite{everingham2010pascal} dataset, we only report the results of AP$_{50}$, denoted as mAP, following \cite{qin2019thundernet,shen2019object}. On the other hand, the widely-used floating-point operations per second (FLOPs), model size (parameters) and frames per second (FPS) are used to measure implementation efficiency.

\subsection{Implementation details}\label{sec:settings}

\subsubsection{Training settings}
DPNet is trained from scratch with a minibatch of 88 images on MS COCO dataset\cite{lin2014microsoft}, using a hardware server platform with a single RTX 2080Ti GPU. The stochastic gradient descent algorithm \cite{bottou2010large} is adopted to train DPNet for 300 epochs, with only 5 epochs warm up. The initialized learning rate is set as $1.5\times10^{-2}$, following cosine learning strategy \cite{loshchilov2016sgdr}, together with weight decay and momentum that are set as $5\times10^{-4}$, and $9\times10^{-1}$, respectively. We also adopt half-precision (FP16) \cite{ge2021yolox} and exponential moving average scheme \cite{micikevicius2017mixed} to reduce GPU memory usage and accelerate training convergence. Instead of using fancy methods \cite{bochkovskiy2020yolov4,zhang2017mixup}, we only employ SSD \cite{liu2016ssd} to perform basic data augmentation. Specifically, we first use color distortion to augment original images, which are further expanded and randomly cropped. After that, transformed images are resized to $320 \times 320$, which are also randomly flipped and normalized. In inference process, following \cite{redmon2018yolov3,wang2021scaled}, we transform DPNet from pytorch to TensorRT FP16 to accelerate detection speed. All settings on Pascal VOC 2007 \cite{everingham2010pascal} are the same with MS COCO \cite{lin2014microsoft}, except the predicted class numbers are 20, instead of 80 categories used in \cite{lin2014microsoft}. Our code is open-source and is publicly available at \url{https://github.com/Huiminshii/DPNet}.

\begin{figure*}[t!] 
	\centering 
	\includegraphics[width=1.0\textwidth, height=0.6\textwidth]{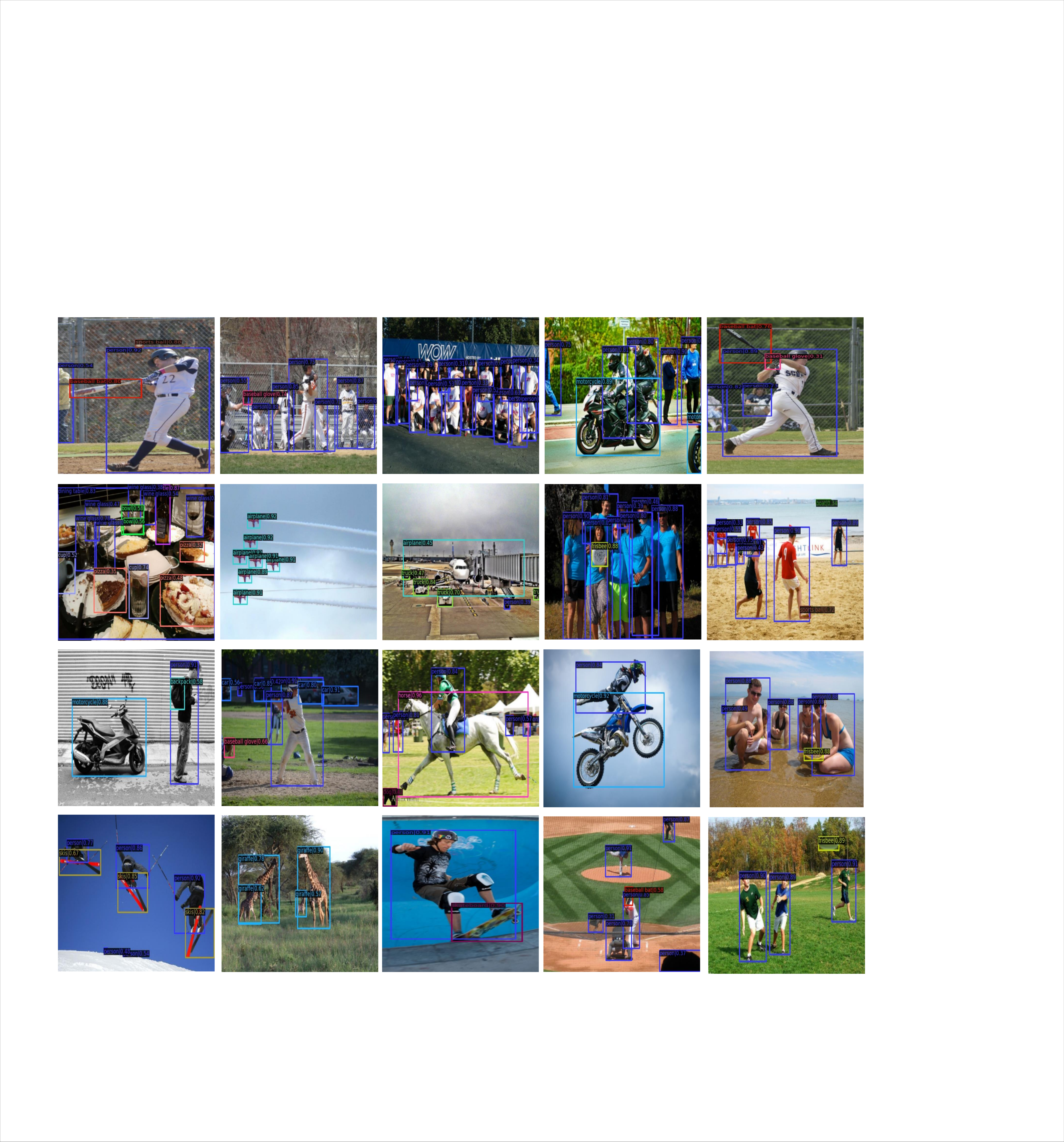} 
	\caption{Some visual examples of qualitative detection results on MS COCO test-dev \cite{lin2014microsoft}. For clarity, the estimated bounding boxes and associated labels are also superimposed on detected objects. (Best viewed in color)} 
	\label{Fig:Vis} 
\end{figure*}

\begin{table*}[t!] 
	\tabcolsep 4.0mm \caption{Comparison with the high-accuracy and real-time object detectors in terms of detection accuracy and implementing efficiency on PASCAL VOC 2007 test set \cite{everingham2010pascal}. ‘-’ denotes the results are not reported. ‘$\dagger$’ and ‘$\ddagger$’ mean DPNet is pre-trained using ImageNet 1K and 21K dataset \cite{deng2009imagenet}, respectively. Note the green and red numbers are with respect to the second-ranked method \cite{qin2019thundernet}.} 
	\begin{center}
		\begin{tabular}{l|c|c|c|c|c|c|c}
			\toprule 
			Method&Year&Backbone&Input Size&FLOPs (G) & Params (M) & $mAP$(\%) & FPS\\
			\midrule
			SSD\cite{liu2016ssd}&ECCV2016&VGG-16& $300\times300$ &35.3&26.29&76.5&46\\
			YOLOV2 \cite{redmon2017yolo9000} &CVPR2017& DarkNet-19 &$416\times416$&17.5&--&76.8&67\\
			\midrule
			Tiny-YOLOv2\cite{redmon2017yolo9000} &CVPR2017&Tiny-DarkNet& $416\times416$& 7.0&60.5&57.1&\textbf{232} \\
			Tiny-YOLOv3\cite{redmon2018yolov3} &Arxiv2017&Tiny-DarkNet& $416\times416$& 5.5&33.2&58.4&210 \\
			Pelee \cite{Pelee} &NeurIPS2018 &PeleeNet&$304\times304$  &1.2 & 6.0 &70.9&125\\
			Tiny-DSOD \cite{li2018tiny}&BMVC2018 &DDB-Net&$300\times300$  &1.1 & -- &72.1& 105\\
			DSOD-Lite\cite{shen2019object} &PAMI2019& DSNet &$300\times300$  & --&10.4&76.7&25.8\\
			ThunderNet \cite{qin2019thundernet}&ICCV2019&SNet-535&$320\times320$&1.3 & -- &78.6& 214\\
			\midrule
			Ours &- &DPNet&$320\times320$&\textbf{1.0}$(\downarrow\textcolor{green}{\textbf{0.1}})$ & \textbf{2.5}$(\downarrow\textcolor{green}{\textbf{3.5}})$ & \textbf{79.2}$(\uparrow\textcolor{red}{\textbf{0.6}})$  & 196\\	
			Ours$^{\dagger}$ &- &DPNet&$320\times320$&\textbf{1.0}$(\downarrow\textcolor{green}{\textbf{0.1}})$ & \textbf{2.5}$(\downarrow\textcolor{green}{\textbf{3.5}})$ & \textbf{80.1}$(\uparrow\textcolor{red}{\textbf{1.5}})$  & 196\\
			Ours$^{\ddagger}$ &- &DPNet&$320\times320$&\textbf{1.0}$(\downarrow\textcolor{green}{\textbf{0.1}})$ & \textbf{2.5}$(\downarrow\textcolor{green}{\textbf{3.5}})$ & \textbf{81.5}$(\uparrow\textcolor{red}{\textbf{2.9}})$  & 196\\	
			\bottomrule
		\end{tabular}\label{tab:det_result_pascal}
	\end{center}
\end{table*}

\subsubsection{Loss settings}
As illustrated in Fig. \ref{Fig:network}, there are three losses used in lightweight detection head: cross-entropy loss $\mathcal{L}_{cls}$ and $\mathcal{L}_{iou}$ for object classification and IoU prediction, respectively, and IoU loss $\mathcal{L}_{reg}$ for bounding box regression. Thus, the total loss $\mathcal{L}$ can be written as:
\begin{equation}\label{eq:total_loss}
	\mathcal{L}= \mathcal{L}_{cls} + \alpha \times \mathcal{L}_{iou}+ \beta \times \mathcal{L}_{reg} 
\end{equation}
Following \cite{ge2021yolox}, two non-negative parameters $\alpha$ and $\beta$ are set defaultedly as 1 and 0.5. To produce ground truth, we employ SimOTA label assignment strategy \cite{ge2021ota,ge2021yolox}.

\subsubsection{Selected State-of-the-art Baselines}
In order to show the advantages of DPNet, we selected 14 state-of-the-art detectors for comparison, including high-accuracy and lightweight networks. The first category contains SSD\cite{liu2016ssd}, RetainNet \cite{lin2017focal}, YOLOs \cite{redmon2018yolov3,wang2021scaled}, ATSS\cite{zhang2020bridging}, Sparse R-CNN\cite{sun2021sparse}, Swin Transformer \cite{liu2021Swin}, and Mobile-Former \cite{chen2021mobile}. On the other hand, the second one includes tiny version of YOLO families \cite{redmon2018yolov3,wang2021scaled}, MobileNet-SSDLite \cite{sandler2018mobilenetv2}, Pelee \cite{Pelee}, Tiny-DSOD \cite{li2018tiny}, MobileDets \cite{xiong2021mobiledets}, ThunderNet \cite{qin2019thundernet}, and Mobile-ViT-SSDLite \cite{mehta2021mobilevit}. Unless special statement, the baseline results are directly borrowed from the corresponding publications.

\subsection{Comparisons with state-of-the-art real-time detectors}
\subsubsection{Experimental Results on MS COCO Dataset}
Table \ref{tab:det_result} reports quantitative comparison results with selected state-of-the-art real-time detectors, demonstrating that DPNet achieves best trade-off in terms of detection accuracy and implementing efficiency. When DPNet is trained from scratch, it achieves 29.6\% AP on MS COCO test-dev, together with only 2.5M model size, 1.04 GFLOPs and 164 FPS. DPNet surpasses all other baselines by large margins in detection AP, AP$_{50}$, and AP$_{75}$ (e.g., it is better by 1.5\%, 0.7\%, and 1.7\% than ThunderNet \cite{qin2019thundernet}, the second-rank real-time detector). Meanwhile, its computational costs are much smaller (e.g., 0.3GFLOPs, 0.4GFLOPs, and 2.4GFLOPs to Pelee \cite{Pelee}, MobileDets \cite{xiong2021mobiledets}, and Tiny-YOLOV4 \cite{wang2021scaled}), and has fewer network size (e.g., 0.2M, 1.8M, and 3.5M to Mobile-VIT \cite{mehta2021mobilevit}, MobileNetV2 \cite{sandler2018mobilenetv2}, and Pelee \cite{Pelee}). Among all baselines, although MobileNetV2 \cite{sandler2018mobilenetv2}, together with SSDLite detection head, saves approximately 20\% GFLOPs of our DPNet, it delivers poor detection results of 8.3\% AP drop with 1.8M heavier model size. To further improve detective accuracy, the backbone of DPNet is also pre-trained using ImageNet 1K and 21K dataset \cite{deng2009imagenet}, respectively, bringing 0.6\% and 0.9\% AP improvement with respect to trained DPNet from scratch.

Table \ref{tab:det_result} also reports the results compared with some high-accuracy detectors that still achieve approximately real-time speed. Although these heavy networks have higher detection performance than our DPNet, they often require dozens even hundreds of GFLOPs and parameters that are unsuitable for real-world applications with limited computational resources and restricted storage memories. Particularly, it is intriguing that DPNet is even superior to YOLOV3 \cite{redmon2018yolov3} that has heavier model size. Mobile-Former\cite{chen2021mobile}, another detector also with dual-path backbone, outperforms DPNet by a margin of 3.7\% AP, yet its GFLOPs are nearly $161\times$ larger than DPNet.

\begin{figure*}[t!] 
	\centering 
	\includegraphics[width=1.0\textwidth, height=0.6\textwidth]{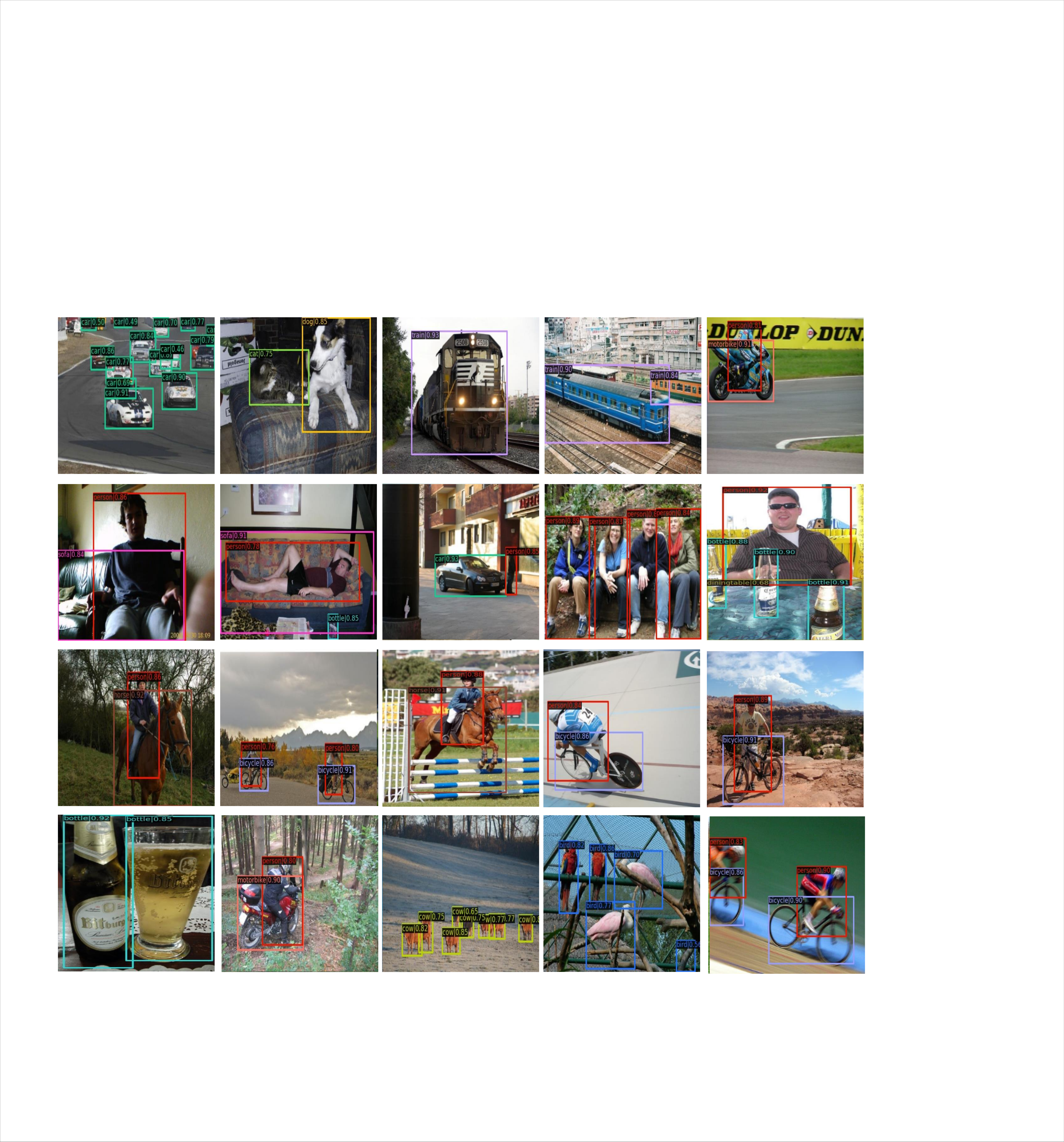} 
	\caption{Visual Examples of qualitative  detection results on Pascal VOC 2007 test set \cite{everingham2010pascal}. For clarity, the estimated bounding boxes and associated labels are also superimposed on detected objects. (Best viewed in color)} 
	\label{Fig:Vis_pascal} 
\end{figure*}

Fig. \ref{Fig:Vis} shows qualitative detective results of some visual examples on MS COCO test-dev. For friendly visualization, the estimated bounding boxes and associated labels are also superimposed on detected objects. It demonstrates that DPNet not only correctly classifies objects within different scales, but also produces accurate bounding boxes for all objects. For instance, ‘‘players’’ in the second and third examples as well as ‘‘people’’ in tenth and eleventh examples are crowded or heavily overlapped, yet DPNet is able to accurately detect them. Moreover, DPNet can tackle object scale variations, such as ‘‘plane’’ and ‘‘truck’’ in eighth example as well as ‘‘giraffe’’ in seventeenth example. Finally, our method shows excellent ability to correctly detect tiny object instances, such as ‘‘plane’’ in the seventh example and tiny ‘‘balls’’ in the first, fifth, tenth and nineteenth examples, respectively. All the results on this dataset show that DPNet learns a powerful representation to capture spatial-wise/channel-wise dependencies and interactions, yielding impressive detection performance with very limited computational overheads.

\subsubsection{Experimental Results on Pascal VOC 2007 Dataset}
In Table \ref{tab:det_result_pascal}, we have also reported the quantitative results on Pascal VOC 2007 test set \cite{everingham2010pascal}, where dual-path backbone is first pre-trained on MS COCO \cite{lin2014microsoft}, and then fine-tuned on the union of Pascal VOC 2007 and 2012. DPNet still yields the best trade-off with 79.2\% mAP, together with only 2.5M model size and 1.0 GFLOPs, outperforming other state-of-the-art real-time detectors by a large margin. For example, compared with second-rank ThunderNet \cite{qin2019thundernet}, involving more complicated backbone that includes hundred layers, DPNet is easier to execute with 0.3GFLOPs drop, yet obtaining a large margin of 0.6\% mAP improvement. To further improve performance, we also utilize ImageNet 1K and 21K \cite{deng2009imagenet} to pre-train dual-path backbone. With the same model size and GFLOPs, mAP averagely increases 2.2\%. Regarding to FPS, DPNet runs faster than most baselines, such as Pelee \cite{Pelee}, Tiny-DSOD \cite{li2018tiny}, and DSOD-Lite \cite{shen2019object}. Although DPNet runs slightly slower than ThunderNet \cite{qin2019thundernet} and Tiny-YOLOV3 \cite{redmon2018yolov3}, we still obtain real-time detective speed of 196FPS, which is enough for real-world applications in timely fashion. Notice that DPNet runs slightly faster than it is evaluated on MS COCO dataset \cite{lin2014microsoft}. This stems form the fact that Pascal VOC 2007 involves less classes for classification and detection. Fig. \ref{Fig:Vis_pascal} illustrates some visual examples of detection results on Pascal VOC 2007 test set. As can be seen, DPNet still obtains visually-pleasing detection results, where large visual variance of object appearance, orientations and scales are well handled, consistenting with the detection outputs shown in Fig. \ref{Fig:Vis}.


\begin{table*}[t!] 
	\tabcolsep 2.2mm \caption{Ablation studies for the contributions of different components in Backbone. The improvement denoted by red numbers are with respect to baseline. Note the number of parameters and GFLOPs are evaluated when input image is of resolution of $320\times320$.} 
	\begin{center}
		\begin{tabular}{ccc|c|c|c|c|c|c|c|c}
			\toprule 
			\multirow{1}{*}{LRP} & \multirow{1}{*}{HRP} & \multirow{1}{*}{Bi-FM}   &\multirow{1}{*}{Params(M)}  & \multirow{1}{*}{Flops(G)}& \multirow{1}{*}{$AP$(\%)}& \multirow{1}{*}{$AP_{50}(\%)$}& \multirow{1}{*}{$AP_{75}(\%)$} & \multirow{1}{*}{$AP_{L}(\%)$} & \multirow{1}{*}{$AP_{M}(\%)$} & \multirow{1}{*}{$AP_{S}(\%)$}\\ 
			\midrule
			\Checkmark &\XSolid &\XSolid   &1.82 &0.87&24.0&38.2&24.6&39.2&23.8&8.2\\
			\Checkmark & \Checkmark&\XSolid   &1.97$(\uparrow\textcolor{red}{\textbf{0.15}})$ &1.11$(\uparrow\textcolor{red}{\textbf{0.24}})$ &26.0$(\uparrow\textcolor{red}{\textbf{2.0}})$ &41.3$(\uparrow\textcolor{red}{\textbf{3.1}})$ &26.6$(\uparrow\textcolor{red}{\textbf{2.0}})$ &41.2$(\uparrow\textcolor{red}{\textbf{2.0}})$ &26.1$(\uparrow\textcolor{red}{\textbf{2.3}})$ &9.8$(\uparrow\textcolor{red}{\textbf{1.6}})$ \\
			\Checkmark &\Checkmark &\Checkmark  &2.16$(\uparrow\textcolor{red}{\textbf{0.34}})$ &1.17$(\uparrow\textcolor{red}{\textbf{0.30}})$ &\textbf{27.0}$(\uparrow\textcolor{red}{\textbf{3.0}})$ &\textbf{42.3}$(\uparrow\textcolor{red}{\textbf{4.1}})$ &\textbf{28.1}$(\uparrow\textcolor{red}{\textbf{3.5}})$ &\textbf{42.6}$(\uparrow\textcolor{red}{\textbf{3.4}})$ &\textbf{28.0}$(\uparrow\textcolor{red}{\textbf{4.2}})$ &\textbf{10.1}$(\uparrow\textcolor{red}{\textbf{1.9}})$\\
			\bottomrule
		\end{tabular}
	\end{center}\label{tab:hrp}
\end{table*}

\subsection{Ablation study}

To understand the underlying behavior of DPNet, this section reports the results of a series of ablation studies. 


\subsubsection{Ablation Study for Components of Backbone}
With fixed neck and detection head, Table \ref{tab:hrp} presents ablation studies that quantify the contributions of different components in backbone, where only LRP is firstly used to build up our baseline, then HRP and Bi-FM are added step-by-step. This experiment shows that each of these components consistently improves the detection performance, together with slight increase of model size and GFLOPs. Among all components, it is observed that HRP brings significantly improvements (e.g., 2.0\%, 3.1\%, and 2.0\% in terms of AP, AP$_{50}$, and AP$_{75}$, respectively), demonstrating the advantage of designing dual-path architecture backbone. In addition, dual-path backbone improves 3.4\% AP$_L$, 4.2\% AP$_M$, and 1.9\% AP$_S$, respectively, mainly benefiting from the HRP that retains fine object details as mush as possible, especially for medium and tiny objects.


\subsubsection{Ablation Study for Different Lightweight Backbones}
Different backbones have been shown to vary in their ability to represent large scale visual data. To further analyze DPNet, we conduct experiments using different lightweight backbones. In particular, also given fixed neck and detection head, we sequentially replace backbone of DPNet with ResNet-18 \cite{resnet}, MobileNetV2 \cite{sandler2018mobilenetv2}, ShuffleNetV2 \cite{ma2018shufflenet}, and Tiny-Darknet \cite{redmon2018yolov3}. As reported in Table \ref{tab:backbone}, ShuffleNetV2 \cite{ma2018shufflenet} has the smallest model size and GFLOPs, yet ranks at the bottom in terms of detection accuracy, probably due to its very limited network capacity. Even though our backbone has nearly 5$\times$ smaller model size and 3.5$\times$ faster running speed, it still achieves better results than ResNet-18 \cite{resnet}. This mainly stems from the fact that the embedded LSCM has powerful ability to capture element-wise interactions with very small computational costs.

\begin{table}[t!] 
	\tabcolsep 0.3mm \caption{Ablation studies using different lightweight backbones. Note the red numbers are with respect to the second-rank method ResNet-18 \cite{resnet}.}
	\begin{center}
		\begin{tabular}{c|c|c|c|c|c}
			\toprule 
			\multirow{1}{*}{Backbone} &\multirow{1}{*}{Params(M)}  & \multirow{1}{*}{Flops(G)}& \multirow{1}{*}{$AP$(\%)}& \multirow{1}{*}{$AP_{50}(\%)$}& \multirow{1}{*}{$AP_{75}(\%)$}\\ 
			\midrule			
			Tiny-DkNet \cite{redmon2018yolov3} &7.33 &1.67 &18.1 &32.6 &17.8   \\		
			ShuNetV2 \cite{ma2018shufflenet} &\textbf{1.57} &\textbf{0.78} &22.5 &37.1 &23.4 \\
			MobNetV2 \cite{sandler2018mobilenetv2} &2.16 &1.17 &26.3 &41.6 &27.3 \\
			ResNet-18 \cite{resnet} &11.98 &4.25 &28.4 &44.6 &30.0 \\
			\midrule
			DPNet  &2.42 &1.20 &\textbf{28.7}$(\uparrow\textcolor{red}{\textbf{0.3}})$ &\textbf{44.8}$(\uparrow\textcolor{red}{\textbf{0.2}})$ &\textbf{30.2}$(\uparrow\textcolor{red}{\textbf{0.2}})$\\
			\bottomrule
		\end{tabular}
	\end{center}\label{tab:backbone}
\end{table}

\begin{table}[t!] 
	\tabcolsep 0.8mm \caption{Ablation studies for LSCM and comparison with other state-of-the-art attention modules. Note the red numbers are with respect to the baseline.}
	\begin{center}
		\begin{tabular}{c|c|c|c|c|c}
			\toprule 
			\multirow{1}{*}{Method} &\multirow{1}{*}{Params(M)}  & \multirow{1}{*}{Flops(G)}& \multirow{1}{*}{$AP$(\%)}& \multirow{1}{*}{$AP_{50}(\%)$}& \multirow{1}{*}{$AP_{75}(\%)$}\\ 
			\midrule
			Baseline &2.16 &1.17&27.0&42.3&28.1 \\
			\midrule
			SE \cite{hu2018squeeze} &2.23 &\textbf{1.18}&27.3&42.7&28.4 \\
			ECA \cite{ecanet} &\textbf{2.17} &\textbf{1.18}& 27.6&43.1& 28.5 \\
			CBAM \cite{woo2018cbam} &2.23 &\textbf{1.18}&27.4&42.6&28.2\\
			SK \cite{li2019selective} &2.47 &1.24&27.8&43.1&28.9\\
			SA  \cite{zhang2021sa}    &\textbf{2.17} &\textbf{1.18}&27.7&43.4&  28.5   \\
			GC \cite{cao2019gcnet} &2.41 &1.19&28.1&43.2&29.3 \\
			\midrule
			LSCM  &2.42 &1.20 &\textbf{28.7}$(\uparrow\textcolor{red}{\textbf{1.7}})$ &\textbf{44.8}$(\uparrow\textcolor{red}{\textbf{2.5}})$ &\textbf{30.2}$(\uparrow\textcolor{red}{\textbf{2.1}})$\\
			\bottomrule
		\end{tabular}
	\end{center}\label{tab:ASU}
\end{table}

\subsubsection{Ablation Study for LSCM}
There exists many attention modules used to capture global context. Therefore, this section compares introduced LSCM with recent state-of-the-art attention blocks. More specifically, the baseline is constructed using entire DPNet, except omitting LSCM in all ASUs of dual-path backbone. Thereafter, LSCM and other attention modules are alternatively inserted in the same place shown in Fig. \ref{Fig:Blocks}(a). The comparison results are reported in Table \ref{tab:ASU}. It shows that LSCM not only outperforms attention blocks that only investigate channel attention (e.g., SE\cite{hu2018squeeze}, ECA \cite{ecanet}, and GC \cite{cao2019gcnet}), but also surpasses counterparts that involve both spatial and channel attention together (e.g., CBAM \cite{woo2018cbam}, SK \cite{li2019selective}, and SA \cite{zhang2021sa}). Compared with baseline, a slight increase of model size and GFLOPs demonstrates that LSCM is a lightweight and efficient module, yet obtains significant improvement of 1.7\% AP, 2.5\% AP$_{50}$, and 2.1\% AP$_{75}$, respectively. Moreover, it is intriguing that LSCM has nearly the same numbers of parameters and GFLOPs with respect to other attention blocks, but achieves more accurate detection results. 


\begin{table}[t!] 
	\tabcolsep 0.5mm \caption{Ablation studies for LCCM and comparison with other state-of-the-art FPNs. Note the red and green numbers are with respect to the second-rank method ASF\cite{guo2020augfpn}.}
	\begin{center}
		\begin{tabular}{c|c|c|c|c|c}
			\toprule 
			\multirow{1}{*}{Method} &\multirow{1}{*}{Params(M)}  & \multirow{1}{*}{Flops(G)}& \multirow{1}{*}{$AP$(\%)}& \multirow{1}{*}{$AP_{50}(\%)$}& \multirow{1}{*}{$AP_{75}(\%)$}\\ 
			\midrule
			FPN\cite{lin2017feature} &2.79 &1.25&28.7&44.8&30.2 \\			
			BFP\cite{pang2019libra} &2.86 &1.22&29.0&45.0&30.4 \\
			PAFPN\cite{tan2020efficientdet} &3.38 &1.36 &29.1 &45.3 &30.4 \\			
			ASF\cite{guo2020augfpn} &3.04 &1.24&29.3&45.5&30.5 \\
			\midrule
			LCCM-TD  &2.39 &1.01 &29.1$(\downarrow\textcolor{green}{\textbf{0.2}})$ &45.5$(\uparrow\textcolor{red}{\textbf{0.0}})$ & 30.3$(\downarrow\textcolor{green}{\textbf{0.2}})$\\
			LCCM-TD-BU  &\textbf{2.42} &\textbf{1.04} & \textbf{30.1}$(\uparrow\textcolor{red}{\textbf{0.8}})$ & \textbf{46.0}$(\uparrow\textcolor{red}{\textbf{0.5}})$ & \textbf{30.9}$(\uparrow\textcolor{red}{\textbf{0.4}})$\\
			\bottomrule
		\end{tabular}
	\end{center}\label{tab:LCCM}
\end{table}

\subsubsection{Ablation Study for LCCM}
As neck plays an essential role for real-time object detection, this section evaluates the effect of introduced LCCM by fixed backbone and detection head. To deeply analyze LCCM, we first only consider LCCM-TD, and then LCCM-BU is sequentially added. The ablation results are reported in Table \ref{tab:LCCM}, together with the comparison of recent widely-used FPNs. When only LCCM-TD is adopted, compared with \cite{guo2020augfpn}, it has slight performance drop (0.2\% of AP and AP$_{75}$), demonstrating that directly fusing features in a top-down manner is not enough to achieve promising results \cite{tan2020efficientdet}. However, when both LCCM-TD and LCCM-BU are utilized, DPNet outperforms \cite{guo2020augfpn} by a large margin. Furthermore, LCCM with LCCM-TD and LCCM-BU combined has the smallest model size and GFLOPs, yet delivers best detection performance. Concretely, LCCM only has 2.42M parameters and 1.2GFLOPs, but yields 1.3\% and 1.1\% AP improvement with respect to FPN \cite{lin2017feature} and BFP \cite{pang2019libra}. It is interesting that PAFPN \cite{tan2020efficientdet} also employs bottom-up and top-down fusion paths, but LCCM still outperforms it in terms of AP, AP$_{50}$, and AP$_{75}$, respectively, with very limited computational costs.


\begin{table}[t!] 
	\tabcolsep 1.1mm \caption{Ablation studies for pooling size $k$ of LSCM.} 
	\begin{center}
		\begin{tabular}{c|c|c|c|c|c}
			\toprule 
			\multirow{1}{*}{Pool Size $k$} &\multirow{1}{*}{Params(M)}  & \multirow{1}{*}{Flops(G)}& \multirow{1}{*}{$AP$(\%)}& \multirow{1}{*}{$AP_{50}(\%)$}& \multirow{1}{*}{$AP_{75}(\%)$}\\ 
			\midrule
			3 &$\textbf{2.78}$ &$\textbf{1.13}$ &28.5 &44.3 &29.9 \\
			5 &2.79 &1.20 &\textbf{28.8} &\textbf{44.8} &\textbf{30.2} \\
			7  &2.79 &1.61 &28.6 &44.6 &30.1 \\
			9  &2.79 &1.80 &28.6 &44.4 &30.0 \\
			11 &2.79 &1.76 &28.2 &44.1 &29.7 \\
			\bottomrule
		\end{tabular}\label{tab:poolsize}
	\end{center}	
\end{table} 

\subsection{Analysis of Parameter Settings}
\subsubsection{Effect of Pooling Size $k$ of LSCM}
The pooling size $k$ determines element numbers used to calculate mutual dependencies, significantly influencing the computational efficiency of LSCM. We thus evaluate the performance variance along with the change of $k$, ranged from 3 to 11 with updated step 2. The results are reported in Table \ref{tab:poolsize}. As can be seen, the rise of pooling size $k$ produces more feature elements involved in correlation calculation, leading to nearly $2\times$ increase of GFLOPs, yet without significant expansion of model parameters. The best result of 28.8\% AP is obtained when $k$ is 5, thus chosen as default setting in DPNet.


\begin{table}[t!] 
	\tabcolsep 1.5mm \caption{Ablation studies for channel compression ratio $r$ of LSCM.} 
	\begin{center}
		\begin{tabular}{c|c|c|c|c|c}
			\toprule 
			\multirow{1}{*}{Ratio $r$} &\multirow{1}{*}{Params(M)}  & \multirow{1}{*}{Flops(G)}& \multirow{1}{*}{$AP$(\%)}& \multirow{1}{*}{$AP_{50}(\%)$}& \multirow{1}{*}{$AP_{75}(\%)$}\\ 
			\midrule
			1 &4.01 &1.43  &\textbf{29.3} &\textbf{45.1} &\textbf{30.2} \\
			2 &3.31 &1.30 &28.8 &45.0 &29.9  \\
			4 &2.96 &1.23 &28.7 &44.7 &30.1  \\
			8 &2.79 &1.20 &29.0 &44.8 &\textbf{30.2} \\
			16 &\textbf{2.70} &\textbf{1.19} &28.5 &44.7 &30.0 \\
			\bottomrule
		\end{tabular}\label{tab:reduceratio}
	\end{center}	
\end{table} 

\subsubsection{Effect of Reduction Ratio $r$ of LSCM}
Besides pooling size $k$, the reduction ratio $r$ is another important factor that controls the capacity and the running speed of LSCM. As a result, we conduct experiments by changing $r$, and report the results in Table \ref{tab:reduceratio}. Note when $r = 1$, our LSCM degenerates to compute dense attention maps similar to self-attention \cite{wang2018non}, leading to the highest detection AP, but at the same time it has the heaviest model size and the largest computational costs. Apart from this, along with the increase of $r$, the model size and GFLOPs gradually decline, but the best AP peaks at $r=8$, which is also opt to default setting in DPNet.


\section{Conclusion remarks and future work}\label{sec:Conclusion}

This paper has presented a dual-path lightweight network, called DPNet, for real-time object detection. The designed dual-path backbone enables us to extract high-level semantics, and at the same time maintain low-level details. Furthermore, two parallel paths are not independent, since the feature exchange enhances information communications between two paths. To improve representation capability of our DPNet, a lightweight attention block, LSCM, is designed in backbone to capture global interactions, with very few computational overheads. We also extend LSCM into LCCM in neck part, where correlated dependencies are well explored between neighboring scale features with different resolutions. We have evaluated our method on two popular object detection datasets: MS COCO and Pascal VOC 2007. The experimental results demonstrate that DPNet achieves state-of-the-art trade-off in terms of detection accuracy and implementation efficiency.

In the future, we are interested in two directions to improve DPNet. As shown in Table \ref{tab:det_result}, there is still a large performance gap between DPNet and high-accuracy detectors, requiring further efforts to improve our model. In addition to achieving superior performance for real-time object detection, we believe that DPNet can be easily used for other visual tasks, such as image classification \cite{resnet} and semantic segmentation \cite{huang2019ccnet,zhu2019asymmetric}.

\bibliographystyle{IEEEtran} 
\bibliography{refs1.bib} 

\end{document}